\documentclass{article}

\usepackage{microtype}
\usepackage{graphicx}
\usepackage{booktabs} 
\usepackage{amsmath}
\usepackage{subfig} 

\usepackage{amsfonts}       
\usepackage{nicefrac}       
\usepackage{microtype}      

\usepackage{amsmath}
\usepackage{amssymb}
\usepackage{natbib}
\usepackage[usenames, dvipsnames]{color}
\usepackage{mdwlist}

\usepackage[skip=1pt]{caption}
\usepackage{hyperref}

\newcommand{\ica}{\hspace{0.25cm}}
\definecolor{purple1}{rgb}{0.54,0,0.54}

\usepackage[accepted]{icml2018}

\usepackage[skip=3pt]{caption}
\usepackage{icml2018}

\usepackage{bm}

\icmltitlerunning{Decomposition of Uncertainty in Bayesian Deep Learning}
\begin{document} 

\twocolumn[

\icmltitle{Decomposition of Uncertainty in Bayesian Deep Learning\\ for Efficient and Risk-sensitive Learning}

\icmlsetsymbol{equal}{*}

\begin{icmlauthorlist}
\icmlauthor{Stefan Depeweg}{sag,tum}
\icmlauthor{Jos\'e Miguel Hern\'andez-Lobato}{cam}
\icmlauthor{Finale Doshi-Velez}{har}
 \icmlauthor{Steffen Udluft}{sag}
\end{icmlauthorlist}

\icmlaffiliation{sag}{Siemens AG}
\icmlaffiliation{tum}{TU Munich}
\icmlaffiliation{cam}{University of Cambridge}
\icmlaffiliation{har}{Harvard University}

\icmlcorrespondingauthor{Stefan Depeweg}{stdepewe@gmail.com}

\vskip 0.3in
]



\printAffiliationsAndNotice{}  

\begin{abstract} 
Bayesian neural networks with latent variables are scalable and
flexible probabilistic models: They account for uncertainty in the estimation
of the network weights and, by making use of latent variables, can capture
complex noise patterns in the data.
We show how to extract and decompose uncertainty into epistemic
and aleatoric components for decision-making purposes.  This allows us to successfully identify informative points for active learning of functions with
heteroscedastic and bimodal noise.  
Using the decomposition we further define a novel risk-sensitive criterion for reinforcement learning to identify policies that balance 
expected cost, model-bias and noise aversion.
\end{abstract}

\section{Introduction}
Many important problems in machine learning require learning functions in
the presence of noise. For
example, in reinforcement learning (RL), the transition dynamics of a system is
often stochastic.  Ideally, a model for these systems should be able to both express such 
randomness  but also to account for the uncertainty in its parameters. 

Bayesian neural networks (BNN) are probabilistic models that place the flexibility of neural networks in a Bayesian framework \citep{blundell2015weight,gal2016uncertainty}.
In particular, recent work has extended BNNs with  latent input variables  (BNN+LV)
to estimate functions with complex stochasticity such as bimodality or heteroscedasticity \citep{depeweg2016learning}.  This model class can describe complex stochastic
patterns via a distribution over the latent input variables (aleatoric uncertainty), while, at the same time, account for model uncertainty via a distribution over weights (epistemic uncertainty).


In this work we show how to  perform and utilize a decomposition of uncertainty
in aleatoric and epistemic components  for decision making purposes. Our contributions 
are:
\vspace{-0.2cm}
\begin{itemize*}
\item  We  derive two decompositions that extract epistemic 
and aleatoric uncertainties from the predictive distribution of BNN+LV (Section \ref{main_idea}).
\item We demonstrate 
that with this uncertainty decomposition, the BNN+LV  identifies informative regions even in bimodal and heteroscedastic cases, enabling efficient active learning in the presence of complex noise (Section \ref{sec:al}).
\item We derive a novel risk-sensitive criterion for  model-based RL based on the uncertainty decomposition, enabling a domain expert to trade-off the risks of reliability, 
which originates from model bias and the risk induced by stochasticity (Section \ref{sec:rl}).
\end{itemize*}
\vspace{-0.2cm}
While using uncertainties over transition probabilities to avoid worst-case behavior has been well-studied  in discrete MDPs (e.g. \citep{shapiro2002minimax,nilim2005robust,bagnell2001solving}, to our knowledge, our work is the first to consider continuous non-linear functions with complex noise.  

\def\H{\mathrm{H}}
\newcommand{\x}{\mathbf{x}}
\newcommand{\EI}{\textrm{EI}}
\newcommand{\C}{\mathcal{C}}
\newcommand{\given}{\,|\,}
\newcommand{\DistGam}{\text{Gam}}
\vspace{-0.1cm}
\section{Background: BNN+LV}\label{sec:bnnswlv}

We review a recent family of probabilistic models for multi-output regression.
These models were previously introduced by \citet{depeweg2016learning},  we
refer to them as Bayesian Neural Networks with latent variables (BNN+LV).

Given a dataset ${\mathcal{D} = \{ \mathbf{x}_n, \mathbf{y}_n \}_{n=1}^N}$, formed
by feature vectors~${\mathbf{x}_n \in \mathbb{R}^D}$ and targets~${\mathbf{y}_n
\in \mathbb{R}}^K$, we assume that~${\mathbf{y}_n =
f(\mathbf{x}_n,z_n;\mathcal{W}) + \bm \epsilon_n}$, where~$f(\cdot ,
\cdot;\mathcal{W})$ is the output of a neural network with weights
$\mathcal{W}$ and $K$ output units. The network receives as input the feature vector $\mathbf{x}_n$ and the
latent variable $z_n \sim \mathcal{N}(0,\gamma)$.
We choose rectifiers, ${\varphi(x) = \max(x,0)}$, as activation functions for the hidden layers and
 the identity function, ${\varphi(x) = x}$, for the output layer.
The network output is
corrupted by the additive noise variable~$\bm \epsilon_n \sim \mathcal{N}(\bm 0,\bm
\Sigma)$ with diagonal covariance matrix $\bm \Sigma$. The role of the latent variable $z_n$ is to
capture unobserved stochastic features that can affect the network's
output in complex ways. Without $z_n$, randomness would only be given by the additive
Gaussian observation noise $\bm \epsilon_n$, which can only describe limited stochastic
patterns. 
The network has~$L$ layers, with~$V_l$ hidden units in layer~$l$,
and~${\mathcal{W} = \{ \mathbf{W}_l \}_{l=1}^L}$ is the collection of~${V_l
\times (V_{l-1}+1)}$ weight matrices.  
The $+1$ is introduced here to account
for the additional per-layer biases. 
We approximate the exact posterior $p(\mathcal{W},\mathbf{z}\given\mathcal{D})$
with: 
\begin{align}
q(\mathcal{W},\mathbf{z}) =  & \underbrace{\left[ \prod_{l=1}^L\! \prod_{i=1}^{V_l}\!  \prod_{j=1}^{V_{l\!-\!1}\!+\!1} \mathcal{N}(w_{ij,l}| m^w_{ij,l},v^w_{ij,l})\right]}_{\text{\small $q(\mathcal{W})$}} \times \nonumber \\
&\underbrace{\left[\prod_{n=1}^N \mathcal{N}(z_n \given m_n^z, v_n^z) \right]}_{\text{\small $q(\mathbf{z})$}}\,.
\label{eq:posterior_approximation}
\end{align}
The parameters~$m^w_{ij,l}$,~$v^w_{ij,l}$ and ~$m^z_n$,~$v^z_n$ are determined
by minimizing a divergence between $p(\mathcal{W},\mathbf{z}\given\mathcal{D})$
and the approximation $q$. The reader is referred to the work of
\citet{hernandez2016black,depeweg2016learning} for more details on this. In
our experiments, we tune $q$ using black-box
$\alpha$-divergence minimization with $\alpha=1.0$. While other values of
$\alpha$ are possible, this specific value produced better uncertainty
decompositions in practice: see Section \ref{sec:al} and the supplementary
material for results with $\alpha=0.5$ and $\alpha = 0$ (variational Bayes).

BNN+LV can capture complex stochastic patterns, while at the same time account
for model uncertainty. They achieve this by jointly learning $q(\mathbf{z})$,
which describes the values of the latent variables that were used to generate
the training data, and $q(\mathcal{W})$, which represents uncertainty about 
model parameters. The result is a flexible Bayesian approach for learning
conditional distributions with complex stochasticity, e.g. bimodal or heteroscedastic noise.

\vspace{-0.25cm}
\section{Uncertainty Decomposition in BNN+LV}\label{main_idea}

Let us assume that the targets are one-dimensional, that is, $K=1$.
The predictive distribution of a BNN+LV for the target variable $y_\star$ associated with the test data point $\mathbf{x}_\star$ is

\vspace{-0.35cm}
{\small
\begin{equation}
p(y_\star|\mathbf{x}_\star) = 
\int p(y_\star|\mathcal{W},\mathbf{x}_\star,z_\star)p(z_\star)q(\mathcal{W})\,dz_\star\,d\mathcal{W} \,.\label{eq:final_predictive_dist}
\end{equation}
}where
$p(y_\star|\mathcal{W},\mathbf{x}_\star,z_\star)=\mathcal{N}(y_\star|f(\mathbf{x}_\star,z_\star;\mathcal{W}),\bm
\Sigma)$ is the likelihood function,
$p(z_\star)=\mathcal{N}(z_\star|0,\gamma)$ is the prior on the latent
variables and $q(\mathcal{W})$ is the approximate posterior for $\mathcal{W}$
given $\mathcal{D}$. In this expression $q(\mathbf{z})$ is not used since
the integration with respect to $z_\star$ must be done using the prior $p(z_\star)$. The reason for this is that
the $y_\star$ associated with $\mathbf{x}_\star$ is unknown and consequently, there is no other evidence on $z_\star$
than the one coming from $p(z_\star)$.

In Eq.~(\ref{eq:final_predictive_dist}), the randomness or uncertainty on
$y_\star$ has its origin in $\mathcal{W} \sim q(\mathcal{W})$, $z_\star \sim
p(z_\star)$ and $\epsilon \sim \mathcal{N}(0,\sigma^2)$. This means
that there are two types of uncertainties entangled in our predictons for
$y_\star$: aleatoric and epistemic
\citep{KIUREGHIAN2009105,kendall2017uncertainties}. The aleatoric uncertainty
originates from the randomness of $z_\star$ and $\epsilon$ and cannot be
reduced by collecting more data. By contrast, the epistemic uncertainty originates
from the randomness of $\mathcal{W}$ and can be reduced by collecting more data, which
will typically shrink the approximate posterior $q(\mathcal{W})$.


Eq.~(\ref{eq:final_predictive_dist}) is the tool to use when making predictions
for $y_\star$. However, there are many settings in which, for decision making
purposes, we may be interested in separating the two forms of uncertainty
present in this distribution. 
We now describe two decompositions , each one differing in the metric used to quantify uncertainty: the
first one is based on the entropy, whereas the second one uses the variance. 

Let $\text{H}(\cdot)$ compute the differential entropy of a probability
distribution.  The total uncertainty present in
Eq.~(\ref{eq:final_predictive_dist}) can then be quantified as
$\text{H}(y_\star|\mathbf{x}_\star)$. Furthermore, assume that we do not
integrate $\mathcal{W}$ out in Eq.~(\ref{eq:final_predictive_dist}) and, instead, we
just condition on a specific value of this variable. The result is then 
$p(y_\star|\mathcal{W},\mathbf{x}_\star) = \int
p(y_\star|\mathcal{W},\mathbf{x}_\star,z_\star) p(z_\star)\, dz_\star$ with
corresponding uncertainty $\text{H}(y_\star|\mathcal{W},\mathbf{x}_\star)$.
The expectation of this quantity under $q(\mathcal{W})$, that is, 
$\mathbf{E}_{q(\mathcal{W})}[ \text{H}(y_\star|\mathcal{W},\mathbf{x}_\star)]$,
can then be used to quantify the overall uncertainty 
in Eq.~(\ref{eq:final_predictive_dist}) coming from $z_\star$ and $\epsilon$.
Therefore, $\mathbf{E}_{q(\mathcal{W})}[ \text{H}(y_\star|\mathcal{W},\mathbf{x}_\star)]$,
measures the aleatoric uncertainty.
We can then quantify the epistemic part of the uncertainty in Eq.~(\ref{eq:final_predictive_dist}) by computing the difference between total and aleatoric uncertainties:
\begin{equation}
 \text{H}[y_\star|\mathbf{x}_\star] - 
\mathbf{E}_{q(\mathcal{W})}[\text{H}(y_\star|\mathcal{W},\mathbf{x}_\star)] = I(y_\star,\mathcal{W})\,,
 \end{equation} 
which is the mutual information between $y_\star$ and $\mathcal{W}$.

Instead of the entropy, we can use the variance as a measure of uncertainty. 
Let $\sigma^2(\cdot)$ compute the variance of a probability distribution. 
The total uncertainty present in Eq.~(\ref{eq:final_predictive_dist}) 
is then $\sigma^2(y_\star|\mathbf{x}_\star)$.
This quantity can then be decomposed using the law of total variance:

\vspace{-0.25cm}
\resizebox{\linewidth}{!}{
  \begin{minipage}{\linewidth}
{\small
\begin{flalign}
\sigma^2(y_\star|\mathbf{x}_\star) &= 
\sigma^2_{q(\mathcal{W})}(\mathbf{E}[y_\star|\mathcal{W},\mathbf{x}_\star]) + 
\mathbf{E}_{ q(\mathcal{W})}[\sigma^2(y_\star|\mathcal{W},\mathbf{x}_\star)]\,. \nonumber 
\end{flalign}
}
\end{minipage}}
\vspace{-0.75cm}

where $\mathbf{E}[y_\star|\mathcal{W},\mathbf{x}_\star]$ and
$\sigma^2[y_\star|\mathcal{W},\mathbf{x}_\star]$ are, respectively, the mean and variance of
$y_\star$ according to $p(y_\star|\mathcal{W},\mathbf{x}_\star)$.
In the expression above,
$\sigma^2_{q(\mathcal{W})}(\mathbf{E}[y_\star|\mathcal{W},\mathbf{x}_\star])$
is the variance of $\mathbf{E}[y_\star|\mathcal{W},\mathbf{x}_\star]$ 
when $\mathcal{W}\sim q(\mathcal{W})$.
This term ignores any contribution to the variance of $y_\star$ from $z_\star$ and $\epsilon$ and only considers
the effect of $\mathcal{W}$.
Therefore, it corresponds to the epistemic uncertainty in Eq.~(\ref{eq:final_predictive_dist}).
By contrast, the term 
$\mathbf{E}_{ q(\mathcal{W})}[\sigma^2(y_\star|\mathcal{W},\mathbf{x}_\star)]$
represents the average value of $\sigma^2(y_\star|\mathcal{W},\mathbf{x}_\star)$
when $\mathcal{W}\sim q(\mathcal{W})$.
This term ignores any contribution to the variance of $y_\star$ from $\mathcal{W}$ and,
therefore, it represents the aleatoric uncertainty in Eq.~(\ref{eq:final_predictive_dist}).

In some cases, working with variances can be undesirable because they have square units.
To avoid this problem, we can work with the square root of the previous terms. For example,
we can represent the total uncertainty using

\vspace{-0.5cm}
\resizebox{\linewidth}{!}{
  \begin{minipage}{\linewidth}
{\small
\begin{flalign}
& \sigma(y_\star|\mathbf{x}_\star) = \nonumber\\
& \quad \quad \left\{\sigma^2_{q(\mathcal{W})}(\mathbf{E}[y_\star|\mathcal{W},\mathbf{x}_\star])
+ \mathbf{E}_{q(\mathcal{W})}[\sigma^2(y_\star|\mathcal{W},\mathbf{x}_\star)]\right\}^\frac{1}{2}\,.\label{eq:std_decomp}
\end{flalign}
}
\end{minipage}}
\vspace{-0.5cm}

\vspace{-0.3cm}
\section{Active Learning with Complex Noise}\label{sec:al}
Active learning is the problem of iteratively collecting data so that the final
gains in predictive performance are as high as possible
\cite{settles2012active}. We consider the case of actively learning arbitrary  non-linear
functions with complex noise.  To do so, we apply the information-theoretic framework
for active learning 
described by \citet{mackay1992information}, which is based on the reduction of entropy in
the model's posterior distribution.  Below, we show that this framework naturally results in the
entropy-based uncertainty decomposition from Section \ref{main_idea}.  Next, we demonstrate how this framework, applied to BNN+LV 
enables data-efficient learning in the presence 
of heteroscedastic and bimodal noise.


Assume a BNN+LV is used to describe a batch of training data 
$\mathcal{D}=\{(\mathbf{x}_1,\mathbf{y}_1),\cdots,(\mathbf{x}_N,\mathbf{y}_N)\}$. The expected reduction in
posterior entropy for $\mathcal{W}$ that would be obtained when 
collecting the unknown target $\mathbf{y}_\star$ for the input $\mathbf{x}_\star$ is
\begin{align}
  \text{H}(\mathcal{W}|\mathcal{D}) - & \mathbf{E}_{\mathbf{y}_\star|\mathbf{x}_\star,\mathcal{D}}\left[\text{H}(\mathcal{W}|\mathcal{D}\cup\{\mathbf{x}_\star,\mathbf{y}_\star\})\right] = I(\mathcal{W},\mathbf{y}_\star)\nonumber \\
  &= \text{H}(\mathbf{y}_\star|\mathbf{x}_\star)-\mathbf{E}_{q(\mathcal{W})}\left[ 
\text{H}(\mathbf{y}_\star|\mathcal{W},\mathbf{x}_\star)\right]\,.
\label{eq:al-objective}
\end{align}
Note that this is the epistemic uncertainty that we introduced in
Section \ref{main_idea}, which has arisen naturally in this setting: the most
informative $\mathbf{x}_\star$ for which to collect $y_\star$ next is the one
for which the epistemic uncertainty in the BNN+LV predictive
distribution is the highest.

The epistemic uncertainty in Eq.~(\ref{eq:al-objective}) can be approximated using
standard entropy estimators, e.g. nearest-neighbor methods
\citep{kozachenko1987sample,kraskov2004estimating,gao2016breaking}. For that,
we repeatedly sample $\mathcal{W}$ and $z_\star$ and do forward passes through
the BNN+LV to sample $\mathbf{y}_\star$.
The resulting samples of $\mathbf{y}_\star$ can then be used to
approximate the respective entropies for each $\mathbf{x}_\star$ using the nearest-neighbor approach:

\vspace{-0.5cm}
{\small
\begin{align}
& \text{H}(\mathbf{y}_\star|\mathbf{x}_\star)-\mathbf{E}_{q(\mathcal{W})}\left[ \text{H}(\mathbf{y}_\star|\mathcal{W},\mathbf{x}_\star)\right]
\nonumber \\
&\approx \hat{\text{H}}(\mathbf{y}_\star^1,\ldots,\mathbf{y}^L_\star) - 
\frac{1}{M} \sum_{i=1}^M \left[\hat{\text{H}}(\mathbf{y}_\star^{1, \mathcal{W}_i},\ldots,\mathbf{y}_\star^{L,\mathcal{W}_i})\right]\,.  
\label{eq:entropy_after}
\end{align}
}
\vspace{-0.5cm}

where $\hat{\text{H}}(\cdot)$ is a nearest-neighbor entropy estimate 
given an empirical sample of points, $\mathbf{y}_\star^1,\ldots,\mathbf{y}_\star^L$ are sampled from
$p(\mathbf{y}_\star|\mathbf{x}_\star)$ according to Eq.~(\ref{eq:final_predictive_dist}),
$\mathcal{W}_1,\ldots,\mathcal{W}_M\sim q(\mathcal{W})$ and
$\mathbf{y}_\star^{1, \mathcal{W}_i},\ldots,\mathbf{y}_\star^{L,\mathcal{W}_i}
\sim
p(y_\star|\mathcal{W}_i,\mathbf{x}_\star)$ for $i=1,\ldots,M$.

There are  alternative ways to estimate the entropy, e.g. with histograms or
using kernel density estimation (KDE) \citep{beirlant1997nonparametric}. We choose
nearest neighbor methods because they tend to work well in low dimensions,
are fast to compute (compared to KDE) and do not require much hyperparameter
tuning (compared to histograms). However, we note that for high-dimensional
problems, estimating entropy is a difficult problem. 
\vspace{-0.25cm}
\subsection{Experiments}

\begin{figure*}[t]
\centering
\subfloat[][]{\includegraphics[scale=0.17]{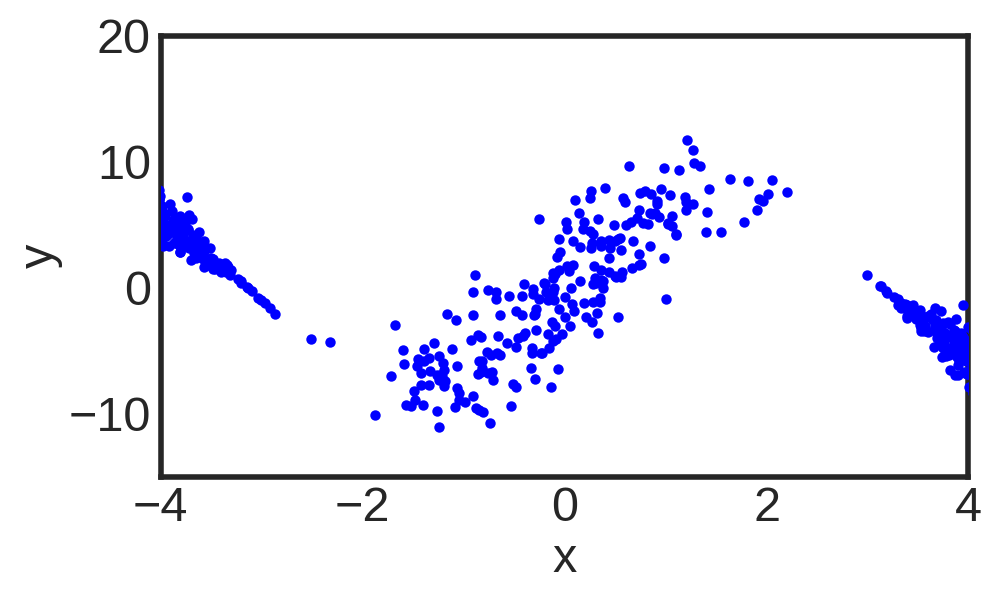}\label{fig:rd}}
\subfloat[][]{\includegraphics[scale=0.17]{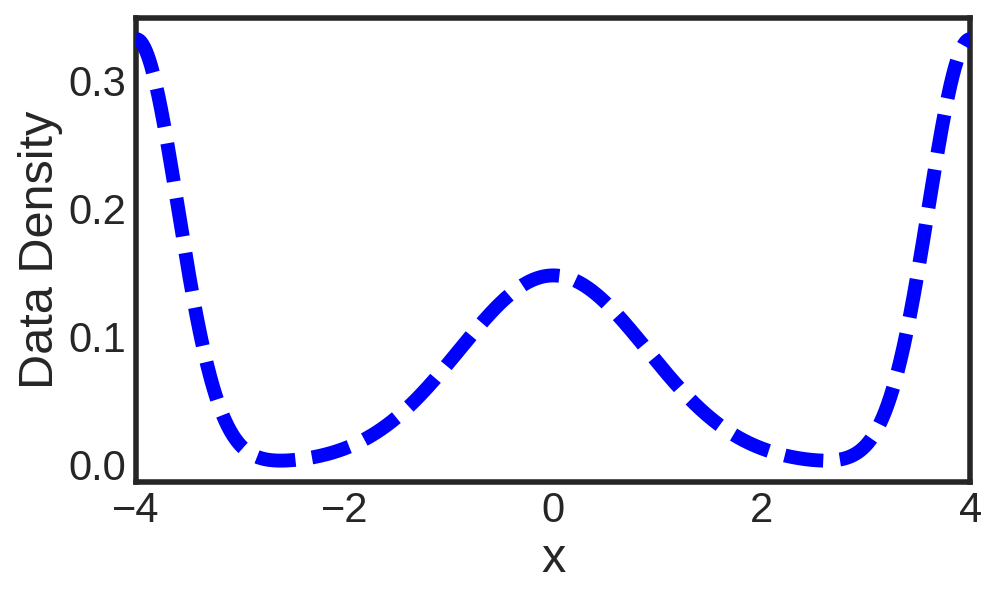}\label{fig:dd}}
\subfloat[][]{\includegraphics[scale=0.17]{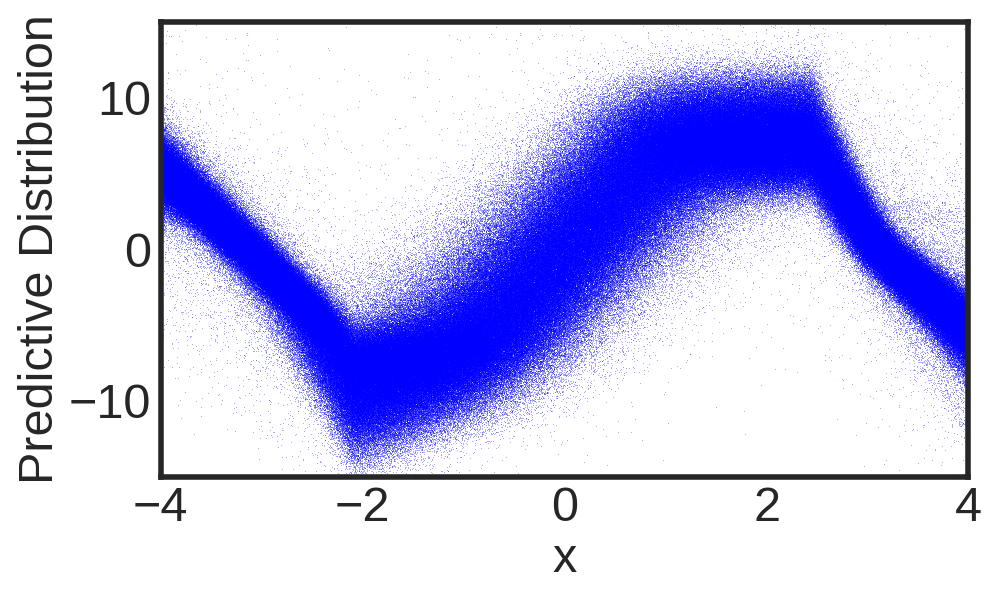}\label{fig:pd}} \\
\vspace{-0.4cm}
\subfloat[][]{\includegraphics[scale=0.17]{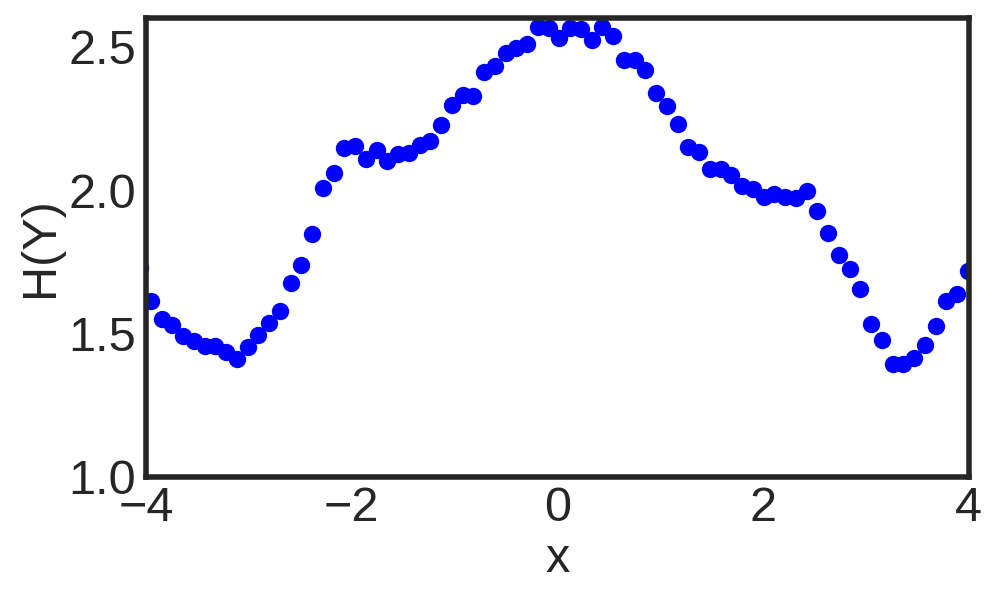}\label{fig:h}}
\subfloat[][]{\includegraphics[scale=0.17]{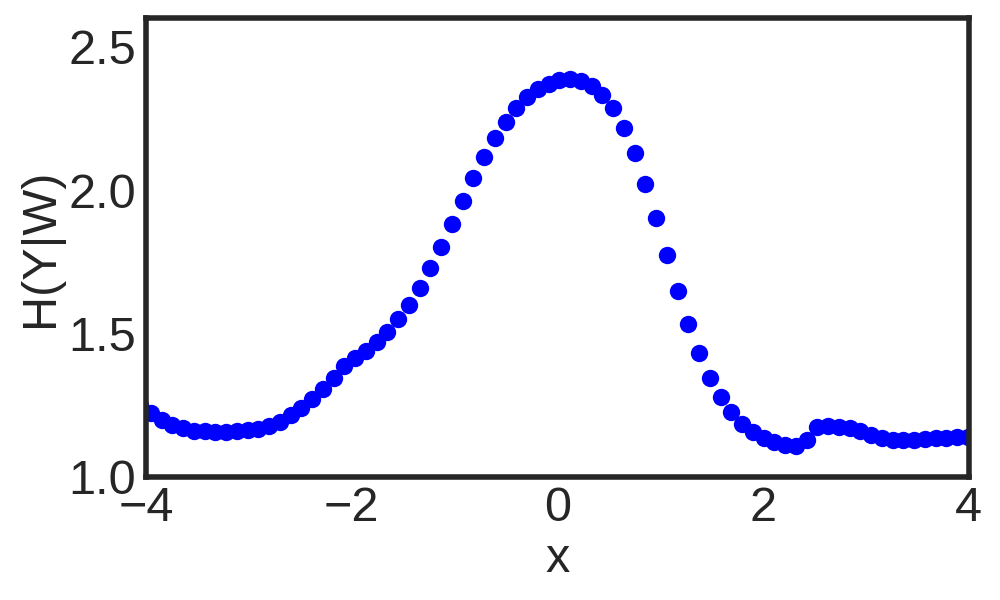}\label{fig:hgw}}
\subfloat[][]{\includegraphics[scale=0.17]{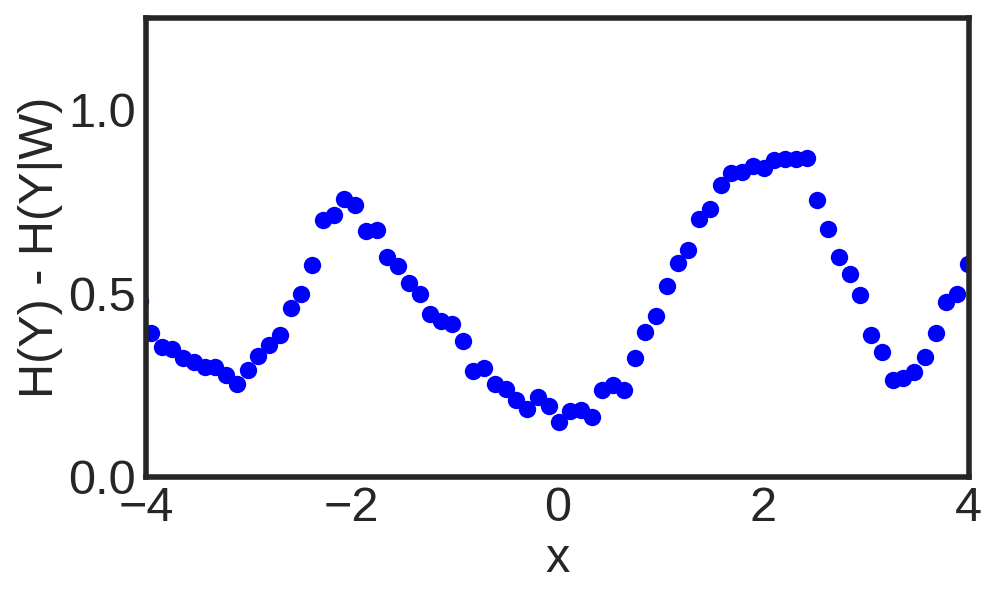}\label{fig:ob}}
\vspace{-0.3cm}
\caption{Uncertainty decomposition on heteroscedastic data. \protect\subref{fig:rd}: Raw data.
  \protect\subref{fig:dd}: Density of $x$ in data. \protect\subref{fig:pd}: Predictive distribution $p(y_\star|x_\star)$.
  \protect\subref{fig:h}: Estimate of $\text{H}(y_\star|x_\star)$. \protect\subref{fig:hgw}: Estimate of $\mathbf{E}_{q(\mathcal{W})} \left[ \text{H}(y_\star|x_\star,\mathcal{W}) \right]$. \protect\subref{fig:ob}: Estimate of entropy reduction $\text{H}(y_\star|x_\star) - \mathbf{E}_{q(\mathcal{W})} \left[ \text{H}(y_\star|x_\star,\mathcal{W}) \right]$.}
  \label{toy_ep}
  \end{figure*}

We  evaluate the active learning procedure on three problems. In
each of them, we first train a BNN+LV with 2 hidden layers and 20 units per
layer. Afterwards, we approximate the epistemic uncertainty as outlined in the
previous section. Hyper-parameter settings and  details for replication can be found in the supplementary material, 
which
includes results for
three other inference methods: Hamiltonian Monte Carlo (HMC), black-box $\alpha$-
divergence minimization with $\alpha=0.5$ and $\alpha = 0$ (variational Bayes).
These results show that the decomposition of uncertainty produced by $\alpha=1$
and the gold standard HMC are similar, but for lower values of $\alpha$ this is not the case.
Our main findings are:
\vspace{-0.25cm}
\paragraph{The decomposition of uncertainty allows us to identify informative inputs when the noise is heteroscedastic.}
We consider a regression problem with heteroscedastic
noise where $y= 7 \sin (x) + 3|\cos (x / 2)| \epsilon$ with $\epsilon \sim
\mathcal{N}(0,1)$. We sample 750 values of the input $x$ from a
mixture of three Gaussians with mean parameters $\{\mu_1= -4,\mu_2=
0,\mu_3= 4\}$, variance parameters
$\{\sigma_1=\frac{2}{5},\sigma_2=0.9,\sigma_3=\frac{2}{5}\}$ and with
each Gaussian component having weight equal to $1/3$ in the mixture.
Figure \ref{fig:rd} shows the data. We have many points at the
borders and in the center, but few in between.

\begin{figure*}[t]
\centering
\subfloat[][]{\includegraphics[scale=0.17]{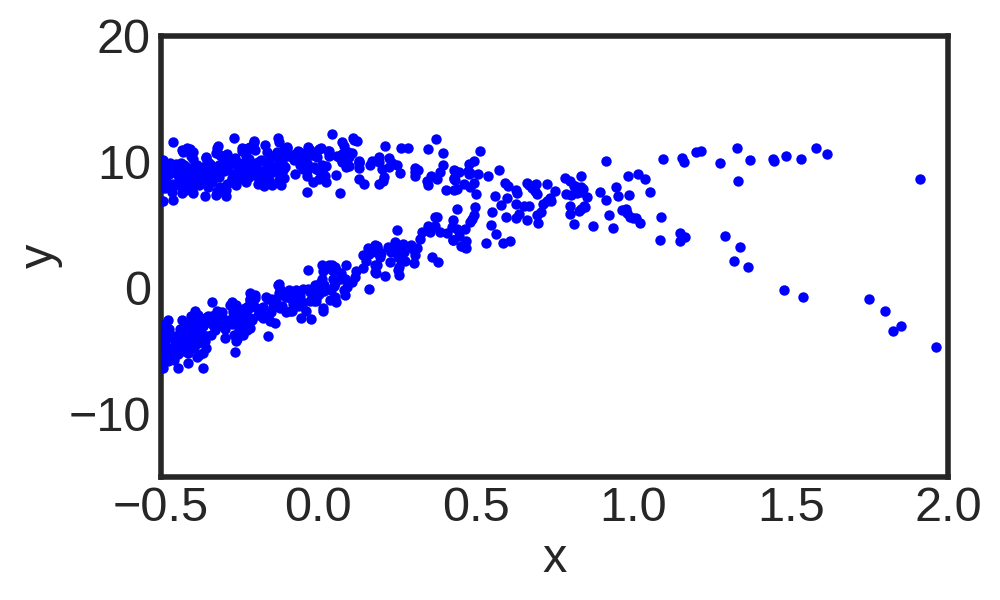}\label{fig:brd}}
\subfloat[][]{\includegraphics[scale=0.17]{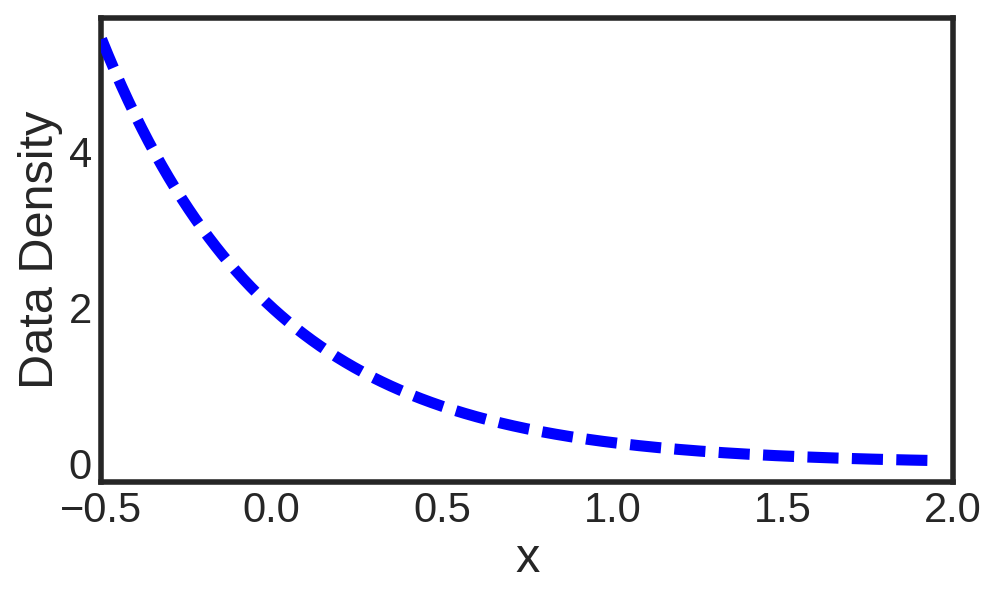}\label{fig:bdd}}
\subfloat[][]{\includegraphics[scale=0.17]{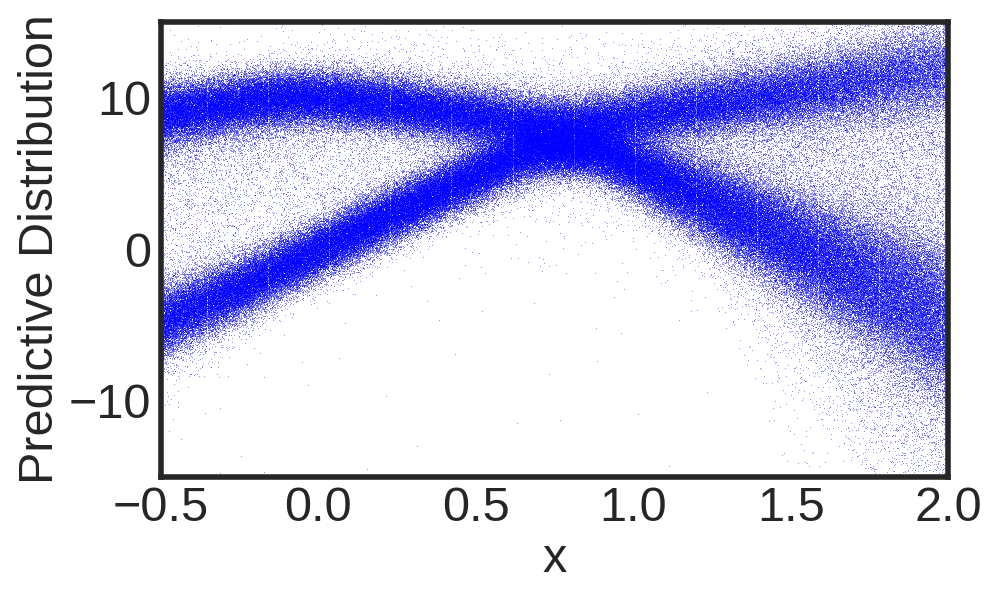}\label{fig:bpd}} \\
\vspace{-0.4cm}
\subfloat[][]{\includegraphics[scale=0.17]{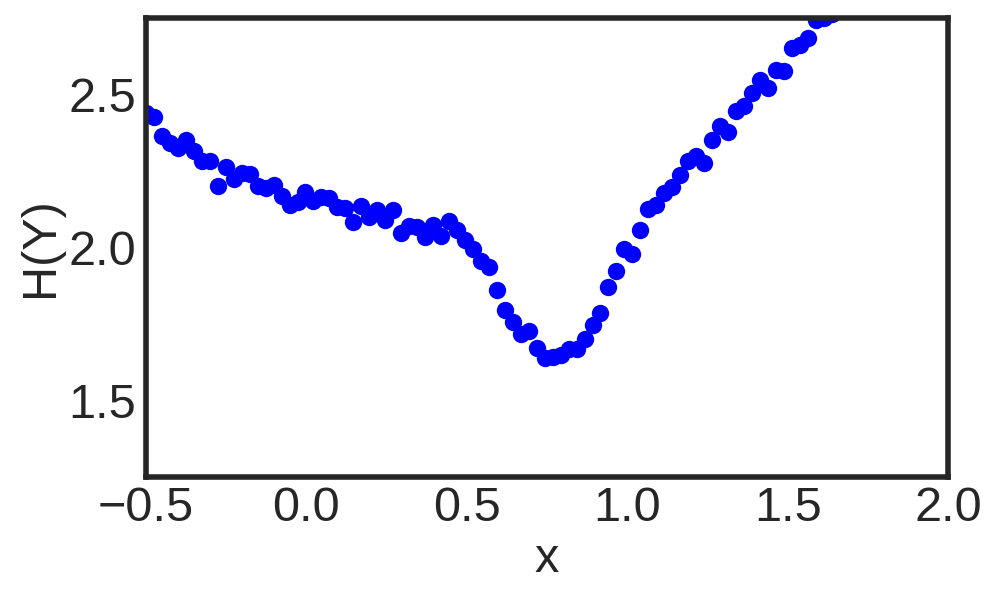}\label{fig:bh}}
\subfloat[][]{\includegraphics[scale=0.17]{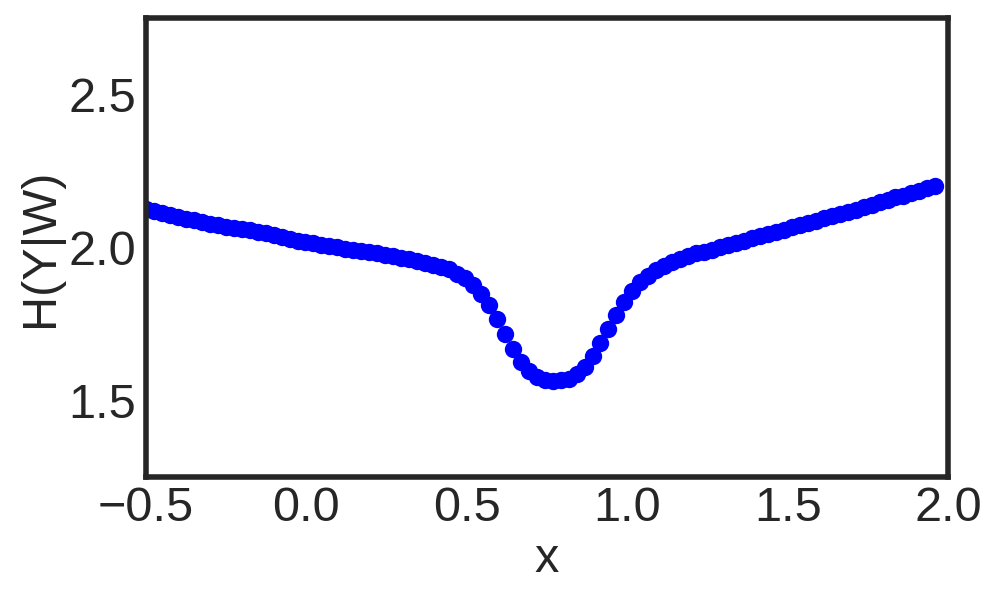}\label{fig:bhgw}}
\subfloat[][]{\includegraphics[scale=0.17]{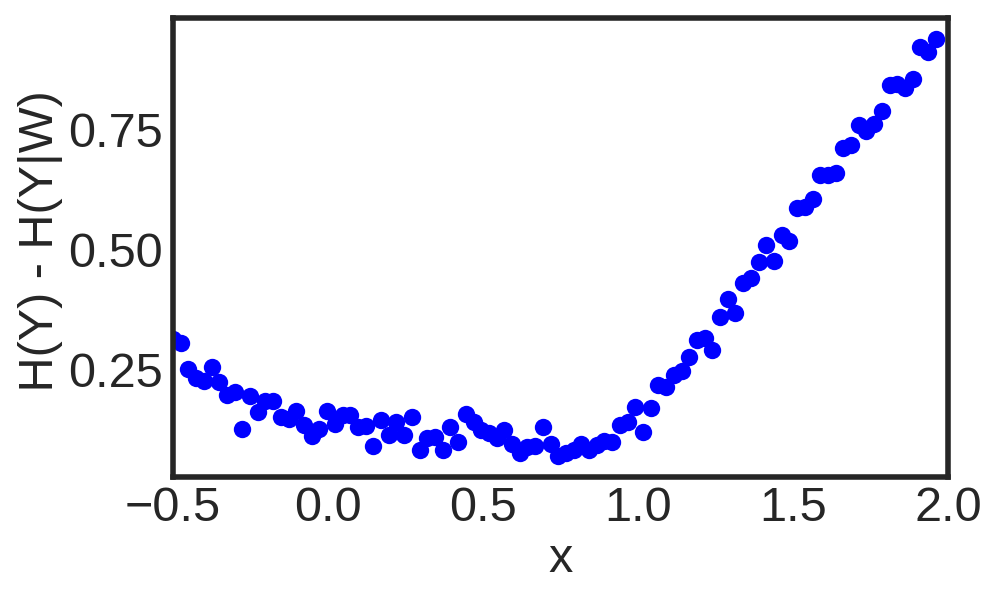}\label{fig:bob}}
\vspace{-0.3cm}
\caption{Uncertainty decomposition on bimodal  data. \protect\subref{fig:rd}: Raw data.
  \protect\subref{fig:dd}: Density of $x$ in data. \protect\subref{fig:pd}: Predictive distribution: $p(y_\star|x_\star)$. \protect\subref{fig:h}: Estimate of $\text{H}(y_\star|x_\star)$. \protect\subref{fig:hgw}: Estimate of $\mathbf{E}_{q(\mathcal{W})} \left[ \text{H}(y_\star|x_\star,\mathcal{W})\right]$. \protect\subref{fig:ob}: Estimate of entropy reduction $\text{H}(y_\star|x_\star) - \mathbf{E}_{q(\mathcal{W})} \left[ \text{H}(y_\star|x_\star,\mathcal{W}) \right]$.}
  \label{toy_bep}
  \end{figure*}
\begin{figure*}[t]
\centering
\subfloat[][]{\includegraphics[scale=0.2]{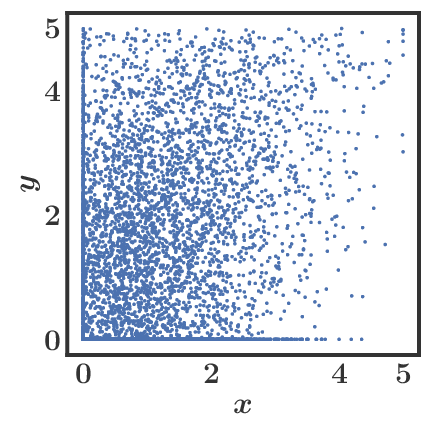}\label{fig:wcrd}}
\subfloat[][]{\includegraphics[scale=0.2]{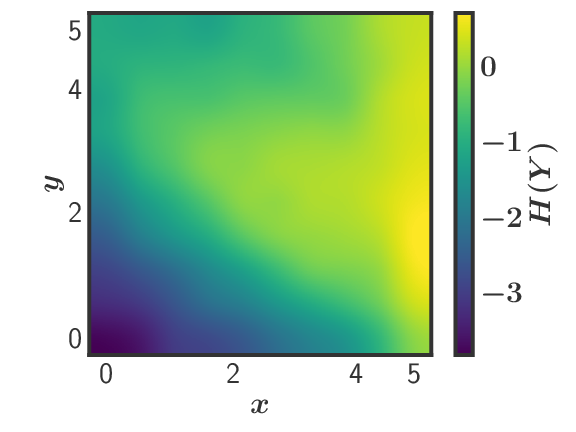}\label{fig:wch}} 
\subfloat[][]{\includegraphics[scale=0.2]{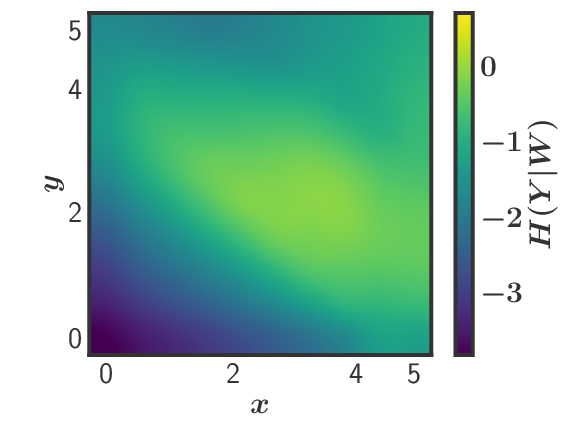}\label{fig:wchgw}}
\subfloat[][]{\includegraphics[scale=0.2]{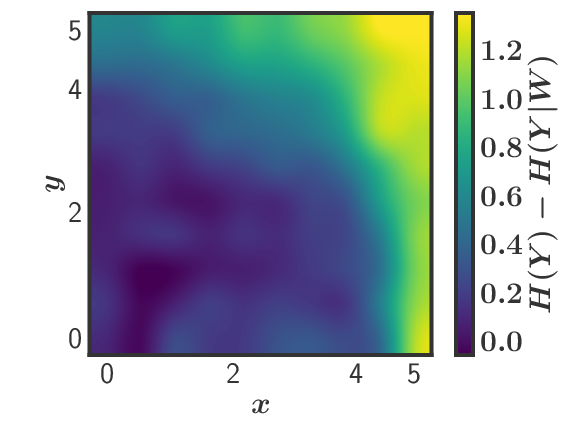}\label{fig:wcob}}
\vspace{-0.3cm}
\caption{Uncertainty decomposition on wet-chicken dynamics. \protect\subref{fig:wcrd}: Raw data.
  \protect\subref{fig:wch}: Entropy estimate $\text{H}(\mathbf{s}_{t+1}|\mathbf{s}_t)$ of predictive distribution for each $\mathbf{s}_t$ (using $\mathbf{a}_t=\{0,0\}$). \protect\subref{fig:hgw}: Conditional entropy estimate $\mathbf{E}_{q(\mathcal{W})} \left[ \text{H}(\mathbf{s}_{t+1}|\mathbf{s}_t,\mathcal{W}) \right]$. \protect\subref{fig:ob}: Estimated entropy reduction.}
  \label{toy_wc}
  \end{figure*}
%

Figure \ref{toy_ep} shows the results obtained (see caption for details). The
resulting decomposition of predictive uncertainty is very accurate: the epistemic uncertainty
(Figure \ref{fig:ob}), is inversely proportional
to the density used to sample the data (Figure \ref{fig:dd}). 
This makes sense, since in this toy problem the most
informative inputs are located in regions where
data is scarce. However, this may not be the case in more complicated settings.
Finally, we note that the total predictive uncertainty (Figure \ref{fig:h})
fails to identify informative regions.
\vspace{-0.5cm}
\paragraph{The decomposition of uncertainty allows us to identify informative
inputs when the noise is bimodal.} Next we consider a toy problem given by a
regression task with bimodal data.  We define $x\in[-0.5, 2]$ and $y=10\sin
(x)+\epsilon$ with probability $0.5$ and $y=10\cos (x)+\epsilon$, otherwise,
where $\epsilon \sim \mathcal{N}(0,1)$ and $\epsilon$ is independent of $x$. We
sample 750 values of $x$ from an exponential distribution with $\lambda=2$.
Figure \ref{fig:brd} shows the data. We have many points on the left, but
few on the right.

Figure \ref{toy_bep} shows the results obtained (see caption for details).
Figure \ref{fig:bpd} shows that the BNN+LV has learned the bimodal structure
in the data and Figure \ref{fig:bh} shows how the total predictive uncertainty
increases on the right, where data is scarce. The aleatoric
uncertainty (Figure \ref{fig:bhgw}), by contrast, has an almost symmetric form around
$x=0.75$, taking lower values at this location. This makes sense since the data
generating process is symmetric around $x=0.75$ and the noise
changes from bimodal to unimodal when one gets closer to $x=0.75$.
Figure \ref{fig:bob} shows an estimate of
the epistemic uncertainty, which as expected increases with $x$.

\vspace{-0.3cm}
\paragraph{The decomposition of uncertainty identifies informative inputs when
noise is both heteroscedastic and bimodal.} We consider data sampled from a 2D
stochastic system called the wet-chicken
\citep{hans2009efficient,depeweg2016learning}, see the supplementary material
for details. The wet-chicken transition dynamics exhibit complex stochastic
patterns: bimodality, heteroscedasticity and truncation (the agent cannot move
beyond the boundaries of state space: $[0,5]^2$). The data are $7,500$ state
transitions collected by random exploration. Figure \ref{fig:wcrd} shows the
states visited. For each transition, the BNN+LV predicts the next state given
the current one and the action applied. Figure \ref{fig:wcob} shows that the
epistemic uncertainty is highest in the top right corner, while data is most
scarce in the bottom right corner. The reason for this result is that the
wet-chicken dynamics bring the agent back to $y=0$ whenever the agent goes
beyond $y=5$, but this does not happen for $y=0$ where the agent just bounces
back. Therefore, learning the dynamics is more difficult and requires more
data at $y=5$ than at $y=0$. The epistemic uncertainty captures this property,
but the total predictive uncertainty (Figure \ref{fig:wch}) does not. 



\begin{table}[t]
\resizebox{\linewidth}{!}{
\begin{tabular}{l@{\ica}r@{$\pm$}l@{\ica}r@{$\pm$}l@{\ica}r@{$\pm$}l@{\ica}}\hline
&\multicolumn{6}{c}{\bf{Method}} \\
\bf{Dataset}&\multicolumn{2}{c}{
$I_\text{bb-$\alpha$}(\mathcal{W}, y_\star)$
}
&\multicolumn{2}{c}{
$\text{H}_{\text{bb-}\alpha}(\mathbf{y}_\star|\mathbf{x}_\star)$
}
&\multicolumn{2}{c}{\textbf{GP}} \\ 
\hline
{\bf Heteroscedastic }  & {\bf -1.79}&{\bf 0.03}&-1.92&0.03&-2.09&0.02 \\ 
{\bf Bimodal} & {\bf -2.04}&{\bf 0.01}&-2.06&0.02&-2.86&0.01 \\
{\bf Wet-chicken}&{\bf 1.18}&{\bf 0.16}&0.57&0.20&-3.22&0.03 \\ 
\hline
\end{tabular}}
\caption{Test log-likelihood in active learning experiments after $150$ iterations.} 
\label{tab:al}
\vspace{-0.3cm}
\end{table}
\vspace{-0.3cm}
\paragraph{Active learning with BBN+LV is improved by using the uncertainty decomposition.}
We evaluate the gains obtained by using Eq.~(\ref{eq:al-objective}) 
with BNN+LV, when collecting data in three toy
active learning problems. We refer to this method as
$I_\text{bb-$\alpha$}(\mathcal{W}, y_\star)$ and compare it with two baselines.
The first one also uses a BNN+LV to describe data, but does not perform a decomposition of predictive
uncertainty, that is, this method uses $\text{H}(\mathbf{y}_\star|\mathbf{x}_\star)$ instead of
Eq.~(\ref{eq:al-objective}) for active learning.
We call this method
$\text{H}_{\text{bb-}\alpha}(\mathbf{y}_\star|\mathbf{x}_\star)$. The second baseline
is given by a Gaussian process (GP) model which collects data according to
$\text{H}(\mathbf{y}_\star|\mathbf{x}_\star)$ since in this case
the uncertainty decomposition is not necessary because the GP model does not include latent variables.
The GP model assumes Gaussian noise and is not able to capture complex stochastic patterns.

The three problems considered correspond to the datasets from Figures
\ref{toy_ep}, \ref{toy_bep} and \ref{toy_wc}. The general set-up is as follows.
We start with the available data shown in the previous figures. At each
iteration, we select a batch of data points to label from a pool set which is
sampled uniformly at random in input space. The selected data is then included
in the training set and the log-likelihood is evaluated on a separate test set.
This process is performed for $150$ iterations and we repeat all experiments 5
times. 

Table \ref{tab:al} shows the results obtained. Overall, BNN+LV outperform GPs
in terms of predictive performance. We also see significant gains of
$I_\text{bb-$\alpha$}(\mathcal{W}, y_\star)$ over
$\text{H}(\mathbf{y}_\star|\mathbf{x}_\star)$ on the heteroscedastic and
wet-chicken tasks, whereas their results are similar on the bimodal task. The
reason for this is that, in the latter task, the epistemic and the total
uncertainty have a similar behaviour as shown in Figures \ref{fig:bh} and
\ref{fig:bob}. Finally, we note that heteroscedastic GPs
\cite{le2005heteroscedastic} will likely perform similar to BNN+LV in the
heteroscedastic task from Figure \ref{toy_bep}, but they will fail in the other
two settings considered (Figures \ref{toy_bep} and \ref{toy_wc}).

\vspace{-0.3cm}
\section{Risk-sensitive Reinforcement Learning}\label{sec:rl}
We  propose an extension of the ``risk-sensitive
criterion" for safe model-based RL \citep{garcia2015comprehensive} to balance 
the risks produced by epistemic and aleatoric uncertainties.  

We focus on batch RL with continuous state and action spaces
\cite{lange2012batch}: we are given a batch of state
transitions $\mathcal{D}=\{(\mathbf{s}_t, \mathbf{a}_t,\mathbf{s}_{t+1})\}$
formed by triples containing the current state $\mathbf{s}_t$, the action
applied $\mathbf{a}_t$ and the next state $\mathbf{s}_{t+1}$. 
In addition to $\mathcal{D}$, we are also given a cost
function $c$. The goal is to obtain from $\mathcal{D}$ a policy in parametric
form that minimizes $c$ on average under the system dynamics.


In model-based RL, the first step consists in learning a dynamics model from $\mathcal{D}$.  
We assume that the true dynamical system can be expressed by an unknown neural network
with latent variables: \begin{equation}\label{eq:transitions} \mathbf{s}_t =
f_\text{true}(\mathbf{s}_{t-1},\mathbf{a}_{t-1},z_t;\mathcal{W}_\text{true})\,,\quad
z_t \sim \mathcal{N}(0,\gamma)\,, \end{equation} where
$\mathcal{W}_\text{true}$ denotes the weights of the network and
$\mathbf{s}_{t-1}$, $\mathbf{a}_{t-1}$ and $z_t$ are the inputs to the network.
We use BNN+LV from Section \ref{sec:bnnswlv} to approximate a posterior $q(\mathcal{W},z)$ using
the batch $\mathcal{D}$ \citep{depeweg2016learning}.

In model-based policy search we optimize a
 policy given by a deterministic neural network with  weights
$\mathcal{W}_{\pi}$.  
This parametric policy returns an action $\mathbf{a}_t$ as a function of $\mathbf{s}_t$, that is,
 $\mathbf{a}_t = \pi(\mathbf{s}_t;\mathcal{W}_\pi)$.
The policy parameters $\mathcal{W}_{\pi}$ can be tuned by minimizing the expectation of the
cost $C = \sum_{t=1}^T c_t$ over a finite horizon $T$ with respect to the
belief $q(\mathcal{W})$, where $c_t = c(\mathbf{s}_t)$. This expected cost is obtained by averaging across
multiple virtual roll-outs as described next. 

Given $\mathbf{s}_0$, we sample
$\mathcal{W}\sim q$ and simulate state trajectories for $T$ steps using
the model $\mathbf{s}_{t+1}=f(\mathbf{s}_t,\mathbf{a}_t,z_t;\mathcal{W})+\bm
\epsilon_{t+1}$ with policy $\mathbf{a}_t = \pi(\mathbf{s}_t;\mathcal{W}_\pi)$,
input noise $z_t \sim \mathcal{N}(0,\gamma)$ and additive noise $\bm
\epsilon_{t+1} \sim \mathcal{N}(\bm 0, \bm \Sigma)$. 
By averaging across these roll-outs, we obtain a Monte Carlo approximation of
the expected cost given the initial state $\mathbf{s}_0$:

\vspace{-0.5cm}
{
\small
\begin{align}
\textstyle J(\mathcal{W}_{\pi})  = \textstyle \mathbf{E}\left[C\right]= \textstyle \mathbf{E}\left[\sum_{t=1}^{T}c_t\right]\,,\label{eq:exact_cost}
\end{align}}
\vspace{-0.5cm}

where, $\mathbf{E}[\cdot]$ denotes here an average across virtual roll-outs
starting from $\mathbf{s}_0$ and, to simplify our notation, we have made the
dependence on $\mathbf{s}_0$ implicit. The policy search algorithm optimizes
the expectation of Eq.~(\ref{eq:exact_cost}) when $\mathbf{s}_0$ is sampled
uniformly from $\mathcal{D}$, that is, $\mathbf{E}_{\mathbf{s}_0 \sim
\mathcal{D}}[ J(\mathcal{W}_{\pi}) ]$.  This quantity can be easily
approximated by Monte Carlo and if model, policy, and cost function are
differentiable, we are able to tune $\mathcal{W}_{\pi}$ by stochastic
gradient descent.


The ``risk-sensitive criterion" \citep{garcia2015comprehensive} 
changes Eq.~(\ref{eq:exact_cost})
to atain a balance between expected cost and risk,
where the risk typically penalizes the deviations of $C$ 
from $\mathbf{E}[C]$ during the virtual roll-outs with initial state $\mathbf{s}_0$.
For example, the risk-sensitive objective could be

\vspace{-0.5cm}
{\small
\begin{align}
\textstyle J(\mathcal{W}_{\pi}) = \textstyle \mathbf{E} \left[C\right] + \beta \mathbf{\sigma} (C)\,,  \label{mf-risk}
\end{align}
}

\vspace{-0.5cm}
where $\mathbf{\sigma}(C)$ is the standard deviation of $C$ across virtual roll-outs starting from $\mathbf{s}_0$ and
the risk-sensitive parameter $\beta$ determines the amount of risk-avoidance ($\beta \ge 0$) or risk-seeking
behavior ($\beta < 0$) when optimizing $\mathcal{W}_{\pi}$.

Instead of working directly with the risk on the final cost $C$, we
consider the sum of risks on the individual costs $c_1,\ldots,c_T$. The reason for this is that
the latter is a more restrictive criterion since low risk on the $c_t$ will imply
low risk on $C$, but not the other way around.
Let $\sigma(c_t)$ denote the standard deviation of $c_t$ over virtual roll-outs starting from $\mathbf{s}_0$.
We can then explicitly write $\sigma(c_t)$ in terms of its aleatoric and epistemic components by using
the decomposition of uncertainty from Eq.~(\ref{eq:std_decomp}). In particular,

\vspace{-0.3cm}
\resizebox{\linewidth}{!}{
  \begin{minipage}{\linewidth}
{\small
\begin{equation}
\sigma(c_t)  = \left\{\sigma^2_{q(\mathcal{W})}(\mathbf{E}[c_t|\mathcal{W}])
+ \mathbf{E}_{q(\mathcal{W})}[\sigma^2(c_t|\mathcal{W})]\right\}^\frac{1}{2} \label{eq:var_decomp}  \,,
\end{equation}
}
\end{minipage}}

\vspace{-0.6cm}
where $\mathbf{E}[c_t|\mathcal{W}]$ and
$\sigma^2(c_t|\mathcal{W})$ denote
the mean and variance of $c_t$
under virtual roll-outs from $\mathbf{s}_0$ performed with policy $\mathcal{W}_{\pi}$ and under 
the dynamics of a BNN+LV with parameters $\mathcal{W}$. 
In a similar manner as in Eq.~(\ref{eq:std_decomp}),
the operators $\mathbf{E}_{q(\mathcal{W})}[\cdot]$
and $\sigma^2_{q(\mathcal{W})}(\cdot)$ in Eq.~(\ref{eq:var_decomp})
compute the mean and variance of their arguments
when $\mathcal{W}\sim q(\mathcal{W})$.

The two terms inside the square root in Eq.~(\ref{eq:var_decomp}) have a clear
interpretation. The first one,
$\mathbf{E}_{q(\mathcal{W})}[\sigma^2(c_t|\mathcal{W})]$, represents the risk
originating from the sampling of $z$ and $\bm \epsilon$ in the virtual
roll-outs. We call this term the aleatoric risk. The second term,
$\sigma^2_{q(\mathcal{W})}(\mathbf{E}[c_t|\mathcal{W}])$, 
encodes the risk originating from the sampling of $\mathcal{W}$
in the virtual roll-outs. We call this term the epistemic risk.






We can now extend the objective in Eq.~(\ref{eq:exact_cost}) with a new risk
term that balances the epistemic and aleatoric risks.  This 
term is obtained by first using risk-sensitive parameters $\beta$ and $\gamma$
to balance the epistemic and aleatoric components in
Eq.~(\ref{eq:var_decomp}), and then summing the resulting expression for
$t=1,\ldots,T$:

\vspace{-0.5cm}
\resizebox{\linewidth}{!}{
\begin{minipage}{\linewidth}
{
\small
\begin{align}
\sigma(\gamma,\beta) =
\sum_{t=1}^T \left\{\beta^2 \sigma^2_{q(\mathcal{W})}(\mathbf{E}[c_t|\mathcal{W}])  + 
\gamma^2 \mathbf{E}_{q(\mathcal{W})}[\sigma^2(c_t|\mathcal{W})]\right\}^\frac{1}{2}\,.\nonumber
\end{align}
}
\end{minipage}
}

\vspace{-0.3cm}
Therefore, our `risk-sensitive criterion" uses the function
$J(\mathcal{W}_{\pi}) = \mathbf{E}\left[ C \right] + \sigma(\gamma,\beta)$, which can be 
approximated via Monte Carlo and optimized 
using stochastic gradient descent.
The Monte Carlo approximation is generated by performing $M \times N$ roll-outs with
starting state $\mathbf{s}_0$ sampled uniformly from $\mathcal{D}$.
For this, $\mathcal{W}$ is sampled from $q(\mathcal{W})$ a total of $M$ times and then, for each of these samples,
$N$ roll-outs are performed with $\mathcal{W}$ fixed and sampling only the
latent variables and the additive Gaussian noise in the BBN+LV. 
Let $z_t^{m,n}$ and $\bm\epsilon_t^{m,n}$ be the samples of the latent variables and the additive Gaussian noise 
at step $t$ during the $n$-th roll-out for the $m$-th sample of 
$\mathcal{W}$, which we denote by $\mathcal{W}^m$. Then
$c_{m,n}(t) = c(\mathbf{s}_{t}^{\mathcal{W}^{m},\{z_{1}^{m,n},\ldots,z_{t}^{m,n}\}, \{\bm{\epsilon}_{1}^{m,n},\ldots,\bm{\epsilon}_{t}^{m,n}\},\mathcal{W}_\pi})$ denotes the cost obtained at time $t$ in that roll-out.
All these cost values obtained at time $t$ are stored in the 
$M\times N$ matrix $\mathbf{C}(t)$. The Monte Carlo estimate of $J(\mathcal{W}_{\pi})$ is then


\vspace{-0.5cm}
\resizebox{\linewidth}{!}{
\begin{minipage}{\linewidth}
{
\small
\begin{align}
 & J(\mathcal{W}_{\pi})  \approx   \sum_{t=1}^{T} \bigg\{ \frac{\mathbf{1}^\text{T} \mathbf{C}(t) \mathbf{1}}{MN} + 
\nonumber \\\displaybreak[0]
    &  \left\{ \beta^2 \hat{\sigma}^2\left[ \mathbf{C}(t)\mathbf{1}/N\right]
 + \gamma^2 \frac{1}{M}\sum_{m=1}^{M} \hat{\sigma}^2\left[ \mathbf{C}(t)_{m,\cdot}\right]\right\}^\frac{1}{2}   \bigg\}\,,
\label{eq:risk_rl1} 
\end{align}
}
\end{minipage}
}

\vspace{-0.3cm}
where $\mathbf{1}$ denotes a vector with all of its entries equal to 1,
$\mathbf{C}(t)_{m,\cdot}$ is a vector with the $m$-th row of $\mathbf{C}(t)$
and $\hat{\sigma}^2[\mathbf{x}]$ returns the empirical variance of the entries in vector $\mathbf{x}$.


By setting $\beta$ and $\gamma$ to specific values in Eq.~(\ref{eq:risk_rl1}), the user can choose different trade-offs between cost,
aleatoric and epistemic risk: for $\gamma=0$ 
the term inside the square root is
$\beta^2$ times $\hat{\sigma}^2\left[ \mathbf{C}(t)\mathbf{1}/N\right]$ which is 
a Monte Carlo  approximation of  the
epistemic risk in Eq.~(\ref{eq:var_decomp}).  Similarly, for $\beta=0$, inside the square root
we obtain
$\gamma^2$ times $\frac{1}{M}\sum_{m=1}^{M} \hat{\sigma}^2\left[ \mathbf{C}(t)_{m,\cdot}\right]$  
which approximates the aleatoric risk. For $\beta=\gamma$ the standard
risk criterion $\sigma(c_t)$ is obtained, weighted by $\beta$.

\vspace{-0.3cm}
\subsection{Model-bias and Noise aversion}

The epistemic risk term in Eq.~(\ref{eq:var_decomp}) can be connected with the concept of
model-bias in model-based RL. A policy with $\mathcal{W}_{\pi}$ is optimized on the model but executed
on the ground truth system. The more  model and ground truth differ, 
the more the policy is 'biased' by the model \citep{deisenroth2011pilco}.
Given the initial state $\mathbf{s}_0$, we can quantify
this bias with respect to the policy parameters $\mathcal{W}_{\pi}$ as

\vspace{-0.3cm}
{
\small
\begin{equation}
b(\mathcal{W}_{\pi}) = \sum_{t=1}^T  \left(\mathbf{E}_\text{true} [c_t] - \mathbf{E}[c_t] \right)^2\,,\label{eq:model_bias}
\end{equation}
}

\vspace{-0.3cm}
where $\mathbf{E}_\text{true} [c_t]$ is the expected cost obtained at time $t$
across roll-outs starting at $\mathbf{s}_0$, under the ground truth dynamics and with policy
$\pi(\mathbf{s}_t ; \mathcal{W}_{\pi})$. $\mathbf{E}[c_t]$ is the same
expectation but under BNN+LV dynamics sampled from $q(\mathcal{W})$ on each
individual roll-out.


Eq.~(\ref{eq:model_bias}) is impossible to compute in practice because we do not know the ground truth dynamics.
However, as indicated in Eq. (\ref{eq:transitions}), we assume that the true dynamic is 
 given by a neural network with latent variables and weights
$\mathcal{W}_\text{true}$. 
We can then rewrite $\mathbf{E}_\text{true} [c_t]$ as
$\mathbf{E}[c_t|\mathcal{W}_\text{true}]$ and
since we do not know $\mathcal{W}_\text{true}$, we can further assume that
$\mathcal{W}_\text{true}\sim q(\mathcal{W})$. The expected model-bias is then

\vspace{-0.2cm}
\begingroup
\allowdisplaybreaks
{
\small
\begin{align}
\mathbf{E}[b(\mathcal{W}_{\pi})] & = \mathbf{E}_{q(\mathcal{W}_\text{true})}
\left\{
\sum_{t=1}^T  \left(\mathbf{E}[c_t|\mathcal{W}_{\text{true}}] - \mathbf{E}[c_t] \right)^2 \right\}\nonumber \\
 &=    \sum_{t=1}^T   \sigma^2_{q(\mathcal{W}_\text{true})}(\mathbf{E}[c_t|\mathcal{W}_\text{true}])\,. \label{eq:risk_rl_sq}
\end{align} 
}
\endgroup
\vspace{-0.1cm}

We see that our definition of epistemic risk also represents an estimate of model-bias in model-based RL.
This risk term will guide the policy to operate in areas of state space where model-bias is expected to be low.

The aleatoric risk term in Eq.~(\ref{eq:var_decomp}) can be connected
with the concept of noise aversion. Let $\sigma^2(c_t|\mathcal{W}_\text{true})$ be the variance obtained at time $t$
across roll-outs starting at $\mathbf{s}_0$, under the ground truth dynamics and with policy
$\pi(\mathbf{s}_t ; \mathcal{W}_{\pi})$.  Assuming
$\mathcal{W}_\text{true}\sim q(\mathcal{W})$,  the expected variance is then 
$\mathbf{E}_{q(\mathcal{W_\text{true}})}[\sigma^2(c_t|\mathcal{W_\text{true}})]$.
This term will guide the policy to
operate in areas of state space where the stochasticity of the cost is low.
Assuming a deterministic cost function this stochasticity is determined by  the model's
predictions that originate from $z_t$ and $\bm\epsilon_t$.

\begin{figure*}[t]
\centering
\subfloat[][]{\includegraphics[scale=0.35]{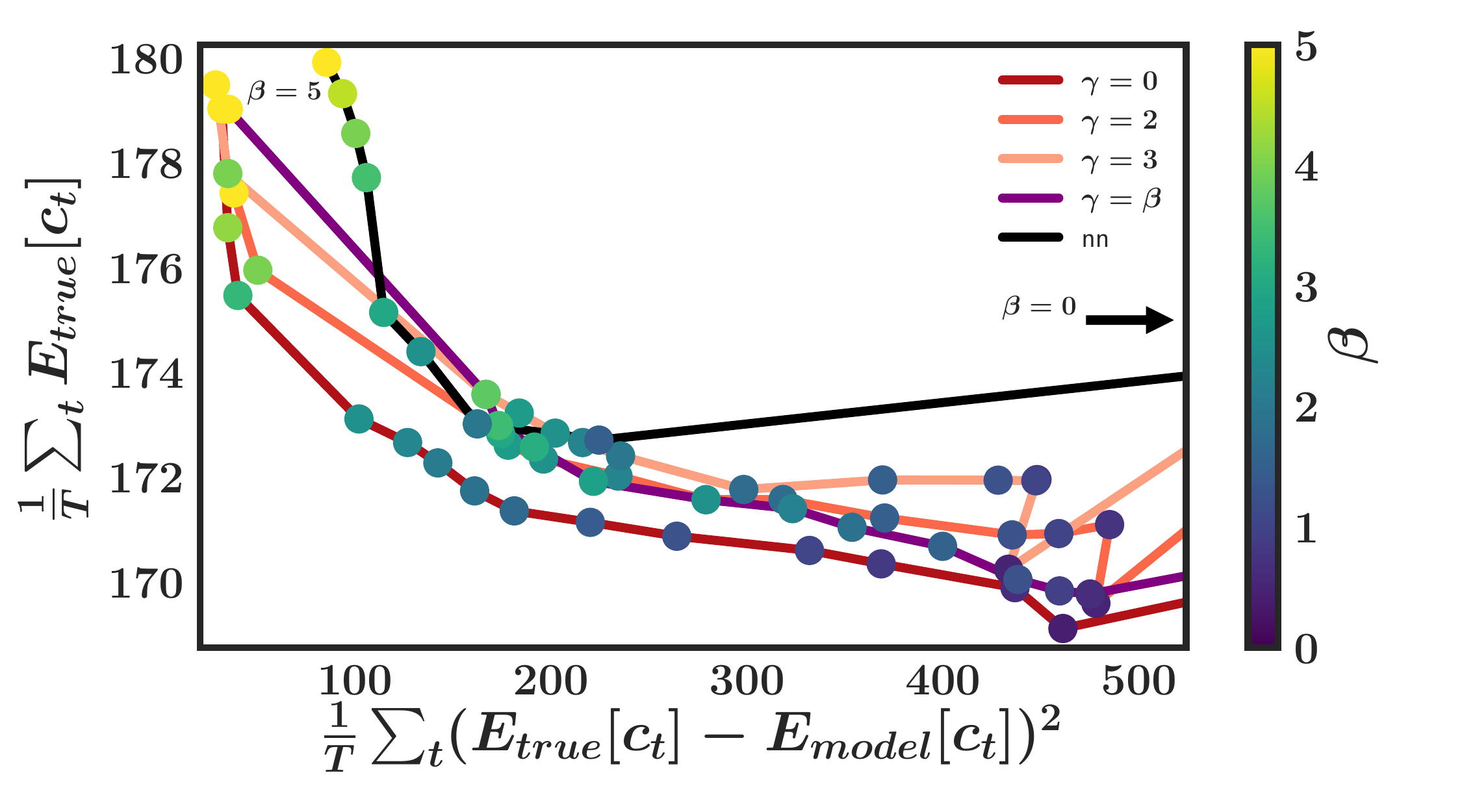}\label{ids_result}}
\subfloat[][]{\includegraphics[scale=0.35]{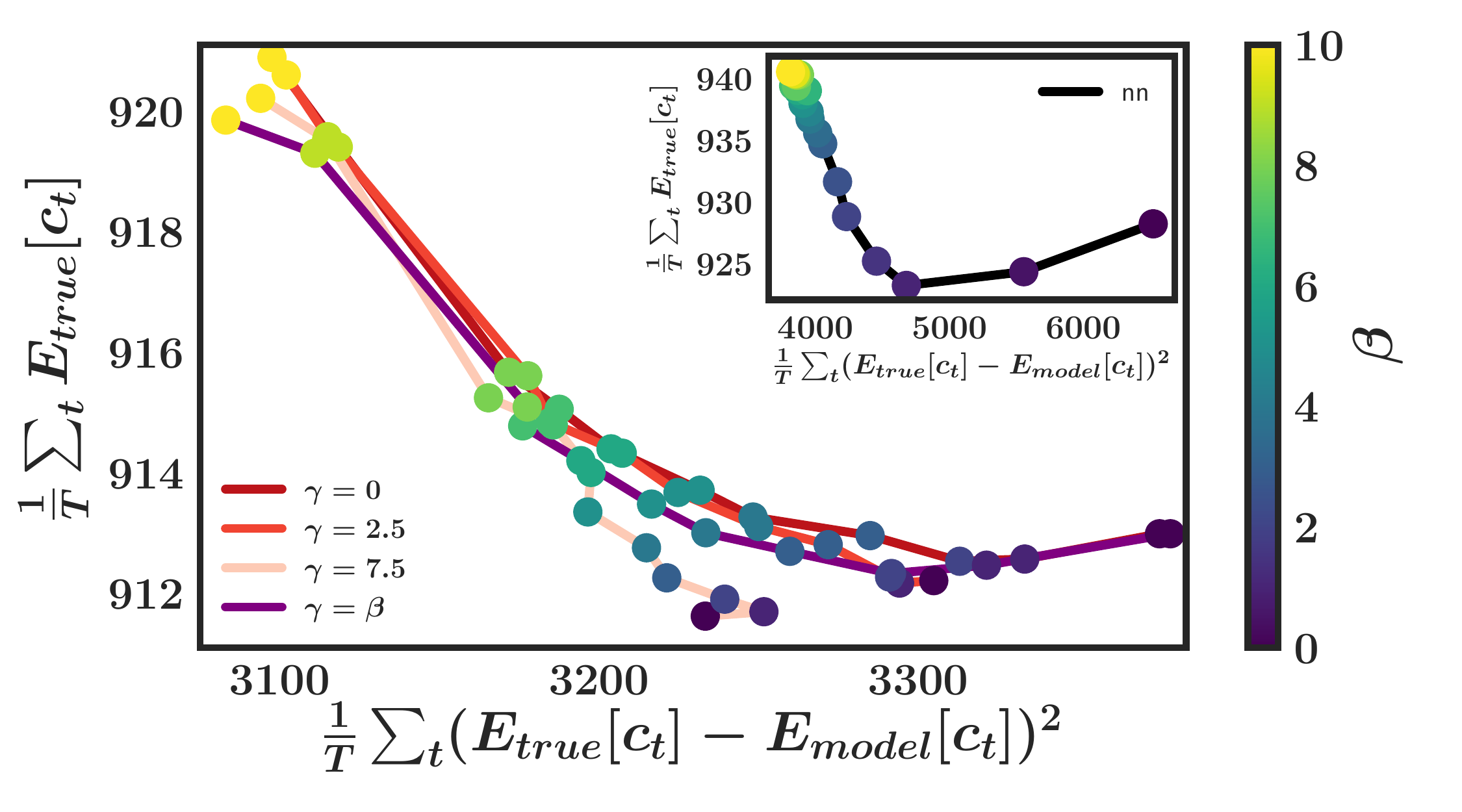}\label{wind_result}}
\vspace{-0.3cm}
\caption{RL experiments. \protect\subref{ids_result}: results on industrial benchmark. \protect\subref{wind_result}: results on
wind turbine simulator. Each curve shows average cost (y-axis) against model-bias (x-axis). Circle color corresponds
to different values of $\beta$ (epistemic risk weight) and curve color indicates different values of $\gamma$ (aleatoric risk weight).
The purple curve is the baseline $\gamma=\beta$. The black curve is nearest neighbor baseline.} 
\end{figure*}

\begin{figure*}[t]
\centering
\subfloat[][]{\includegraphics[scale=0.3]{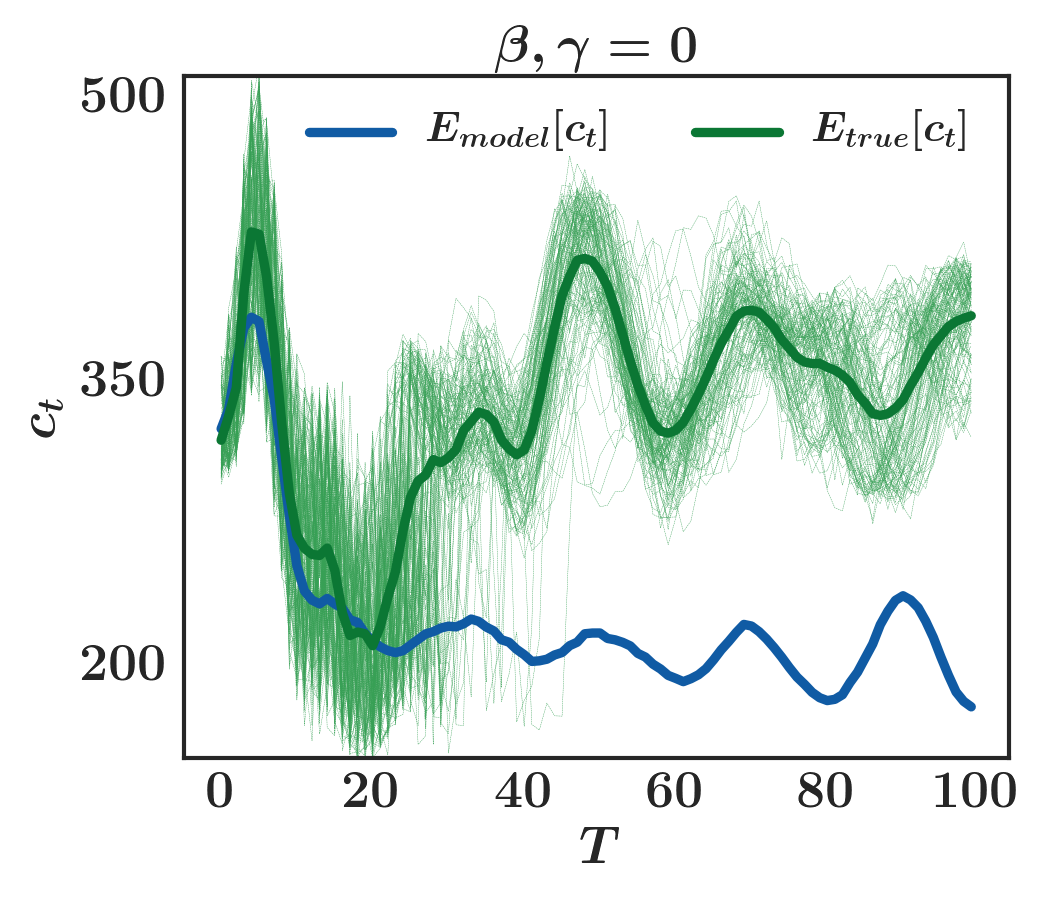}\label{fig:ids_rollout_1}}
\subfloat[][]{\includegraphics[scale=0.3]{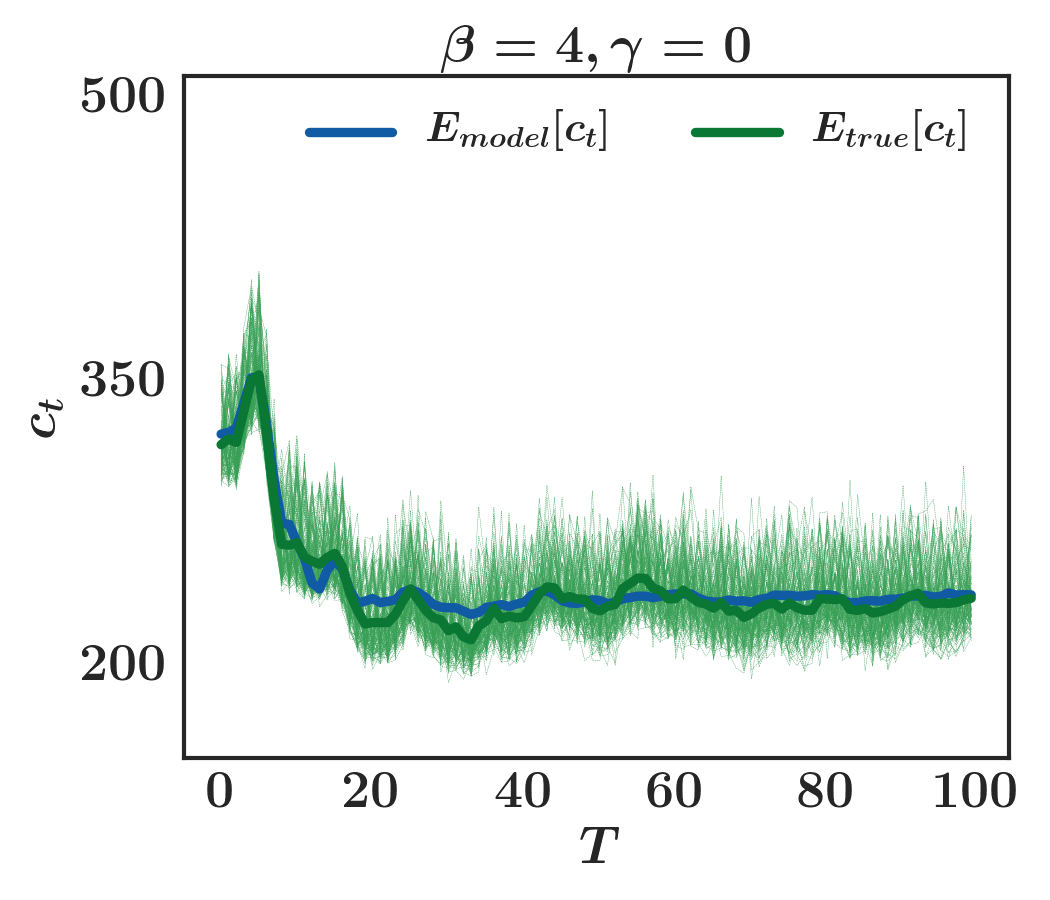}\label{fig:ids_rollout_2}}
\subfloat[][]{\includegraphics[scale=0.3]{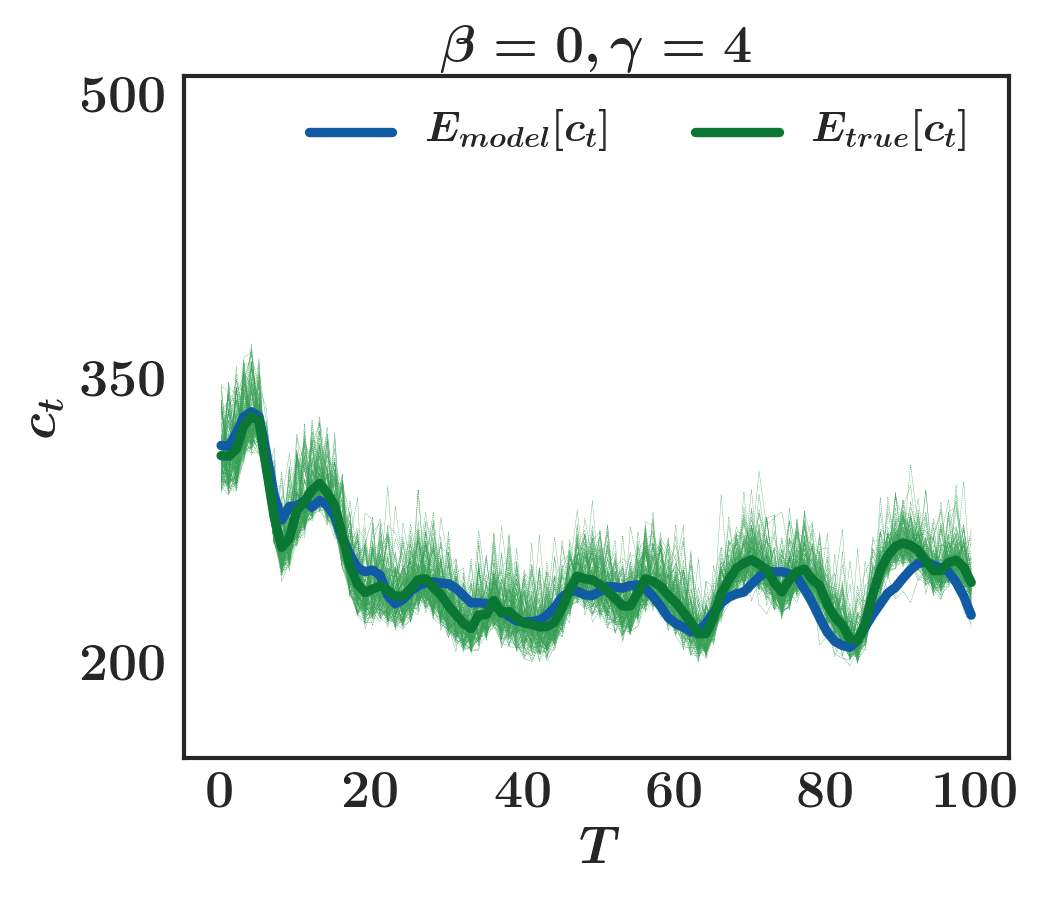}\label{fig:ids_rollout_3}}
\vspace{-0.3cm}
\caption{100 roll-outs on the industrial benchmark ground truth system (light green), their average cost (dark green), 
and the average cost of corresponding roll-outs on the BNN+LV model (blue) for a fixed value of the initial state 
$\mathbf{s}_0$. We show results for three policies with different epistemic  and aleatoric risk trade-offs.
Policies are optimized  using \protect\subref{fig:ids_rollout_1}: no risk
penalty ($\beta,\gamma=0$). \protect\subref{fig:ids_rollout_2}:   
a penalty on the epistemic risk only ($\gamma=0,\beta = 4$).  
\protect\subref{fig:ids_rollout_3}: a penalty on the aleatoric risk only ($\gamma=4,\beta=0$).}
\label{ids_rollout}
\end{figure*}
\vspace{-0.3cm}
\subsection{Experiments} \label{sec:rl_experiments}

We investigate the following questions: To what extent does our new risk
criterion reduce model-bias? What trade-offs do we  observe between
average cost and model-bias?  How does the decomposition compare to other
simple methods?

We consider two model-based RL scenarios. The first one is given by
the industrial benchmark (IB), a publicly available simulator with properties
inspired by real industrial systems \citep{hein2017benchmark}. The second RL
scenario is a modified version of the HAWC2 wind turbine simulator
\citep{larsen20072}, which is widely used for the study of wind turbine
dynamics \citep{larsen2015comparison}.

We are given a batch of data formed by
state transitions generated by a behavior policy $\pi_b$, for example, from an
already running system. The behavioral policy has limited randomness and will
keep the system dynamics constrained to a reduced manifold in state space.
This means that large portions of state space will be unexplored and
uncertainty will be high in those regions. The supplementary material contains
full details on the experimental setup and the $\pi_b$.

We consider the risk-sensitive criterion from Section \ref{sec:rl} for
different choices of $\beta$ and $\gamma$, comparing it with 3 baselines. The
first baseline is obtained by setting $\beta=\gamma=0$. In this case, the policy
optimization ignores any risk. The second baseline is obtained when
$\beta=\gamma$. In this case, the risk criterion simplifies to $\beta
\sigma(c_t)$, which corresponds to the traditional risk-sensitive approach in
Eq.~(\ref{mf-risk}), but applied to the individual costs $c_1,\ldots,c_T$. The
last baseline uses a deterministic neural network to model the dynamics and a
nearest neighbor approach to quantify risk: for each state $\mathbf{s}_t$ generated in a
roll-out, we calculate the Euclidean distance of that state to the nearest
one in the training data. The average value of the distance metric for $\mathbf{s}_t$ across
roll-outs is then an approximation to $\sigma(c_t)$.
To reduce computational cost, we summarize the training data using the centroids
returned by an execution of the k-means clustering method.
We denote this method as the nn-baseline.

Figure \ref{ids_result} shows result on the industrial benchmark. The
$y$-axis in the plot is the average total cost at horizon $T$ obtained by the
policy in the ground truth system. The $x$-axis is the average model-bias in
the ground truth system according to Eq.~(\ref{eq:model_bias}). Each individual
curve in the plot is obtained by fixing $\gamma$ to a specific value (line
colour) and then changing $\beta$ (circle colour). The policy that ignores risk
($\beta = \gamma = 0$) results in both high model-bias and high cost when
evaluated on the ground truth, which indicates overfitting. As $\beta$
increases, the policies put more emphasis on avoiding model-bias, but at the
same time the average cost increases.  The best tradeoff is obtained by the
dark red curve with $\gamma=0$. The risk criterion is then $\beta
\sum_{t=1}^T\sigma_{q(\mathcal{W})}(\mathbf{E}[c_t|\mathcal{W}])$.  In this
problem, adding aleatoric risk by setting $\gamma > 0$ decreases performance.  The
nn-baseline shows a similar pattern as the BNN+LV approach, but the
trade-off between model-bias and cost is worse.

Figure \ref{ids_rollout} shows roll-outs for three different policies and a
fixed initial state $\mathbf{s}_0$. 
Figure \ref{fig:ids_rollout_1} shows
results for a policy learned with $\gamma = \beta = 0$. This policy ignores
risk, and as a consequence, the mismatch between predicted performance on the
model and on the ground truth increases after $t=20$. This result illustrates
how model-bias can lead to policies with high costs at test time.  Figure
\ref{fig:ids_rollout_2} shows results for policy that was trained while
penalizing epistemic risk ($\beta = 4$, $\gamma = 0$). In this case, the
average costs under the BNN+LV model and the ground truth are similar, and the
overall ground truth cost is lower than in Figure \ref{fig:ids_rollout_1}.
Finally, Figure \ref{fig:ids_rollout_3} shows results for a noise averse policy
($\beta = 0$, $\gamma = 4$).  In this case, the model bias is slightly higher
than in the previous figure, but the stochasticity is lower.

The results for wind turbine simulator can be found in Figure
\ref{wind_result}. In this case, the best trade-offs between expected cost and
model-bias are obtained by the policies with $\gamma = 7.5$. These policies are
noise averse and will try to avoid noisy regions in state space. This makes
sense because in wind turbines, high noise regions in state space are those
where the effect of wind turbulence will have a strong impact on the average cost.

\section{Related Work}

The distinction between aleatoric and epistemic uncertainty has been
recognized in many fields within machine learning, often within the
context of specific subfields, models, and objectives. 
\citet{kendall2017uncertainties} consider a decomposition of
uncertainty in the context of computer vision with
heteroscedastic Gaussian output noise, while \citet{mcallister2016bayesian}
consider a decomposition in GPs for model-based RL. 


Within reinforcement learning, Bayesian notions of model uncertainty
have a long history
\citep{urgen1991possibility,schmidhuber1991curious,dearden1999model,
  still2012information,sun2011planning,maddison2017particle}. The
mentioned works typically consider the online case, where model uncertainty
(a.k.a. curiosity) is used to guide exploration, e.g. in
\citet{houthooft2016vime} the uncertainty of a BNN model is used to
guide exploration assuming deterministic dynamics.  In contrast, we
focus on the batch setting with stochastic dynamics.

Model uncertainty is   used in safe or risk-sensitive RL \citep{mihatsch2002risk,garcia2015comprehensive}. In
safe RL numerous other approaches exists for safe exploration \citep{joseph2013reinforcement,hans2008safe,garcia2012safe,berkenkamp2017safe}. 
 Uncertainties over transition probabilities have been studied  in discrete MDPs since a long time \citep{shapiro2002minimax,nilim2005robust,bagnell2001solving} often with a focus on worst-case avoidance. Our
 work extends this to  continuous state and action space using scalable 
 probabilistic models. Our decomposition  enables a practitioner to adjust the optimization criterion
 to specific decision making,.

Within active learning many approaches exist that follow an
information theoretic approach \citep{mackay1992information,hernandez2015probabilistic,guo2007optimistic}.
To our knowledge, all of these approaches however use deterministic methods (mostly GPs) as model class. Perhaps closest to our work is BALD \cite{houlsby2012collaborative}, however because GPs are used, 
this approach cannot model problems with complex noise.

\vspace{-0.3cm}
\section{Conclusion}


We have described a decomposition of predictive uncertainty into its epistemic
and aleatoric components when working with Bayesian neural networks with latent
variables. We have shown how this decomposition of uncertainty can be used for
active learning, where it naturally arises from an information-theoretic
perspective. We have also used the decomposition to propose a novel risk-
sensitive criterion for model-based reinforcement learning which decomposes
risk into  model-bias and noise aversion components.  Our experiments
illustrate how the described decomposition of uncertainty is useful for
efficient and risk-sensitive learning.

{\small
\bibliography{references}
\bibliographystyle{icml2018}
}

\appendix
\onecolumn

\section{Hamiltonian Monte Carlo Solution to Toy Problems}
The goal here is to show that the decomposition we obtained in the active learning examples is not a result
of our approximation using black-box $\alpha$-divergence minimization but a property of BNN+LV themselves. To that end we will approximate using HMC.  After a burn-in of $500,000$ samples we sample from the posterior $200,000$ samples.  We thin out 90\% by only keeping every tenth sample. 
 
\subsection{Heteroscedastic Problem}
\begin{figure*}[h!]
\centering
\subfloat[][]{\includegraphics[scale=0.21]{figures/plots_hetero_new/hetero_1.png}\label{fig:hmcrd}}
\subfloat[][]{\includegraphics[scale=0.21]{figures/plots_hetero_new/hetero_1_5.png}\label{fig:hmcdd}}
\subfloat[][]{\includegraphics[scale=0.21]{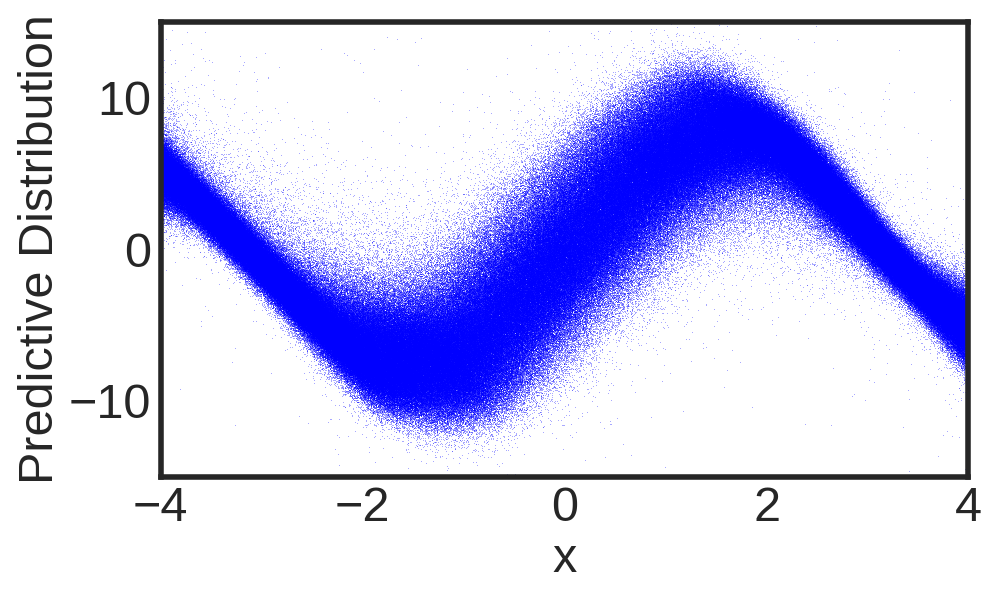}\label{fig:hmcpd}} \\
\vspace{-0.25cm}
\subfloat[][]{\includegraphics[scale=0.21]{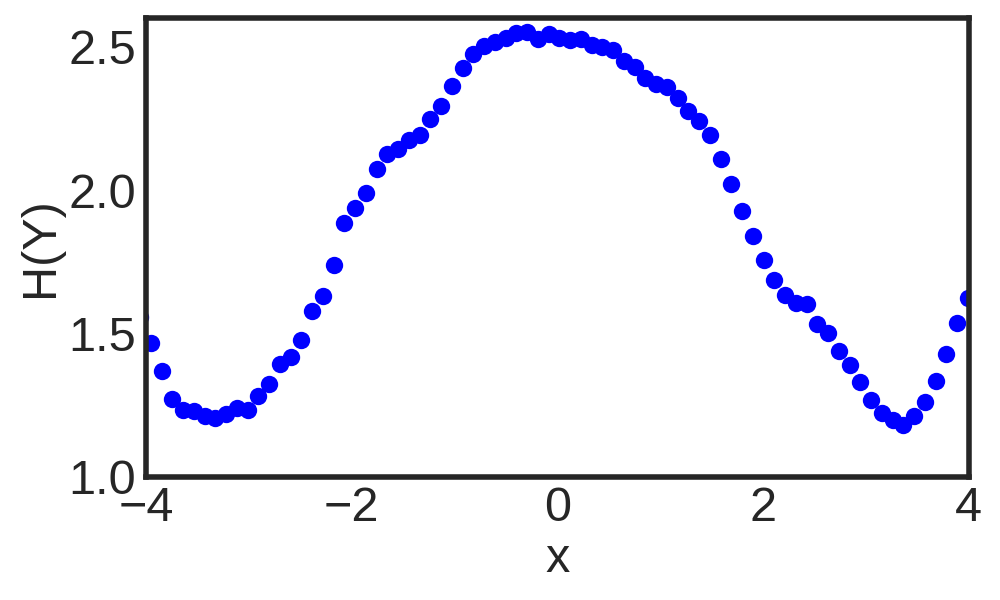}\label{fig:hmch}}
\subfloat[][]{\includegraphics[scale=0.21]{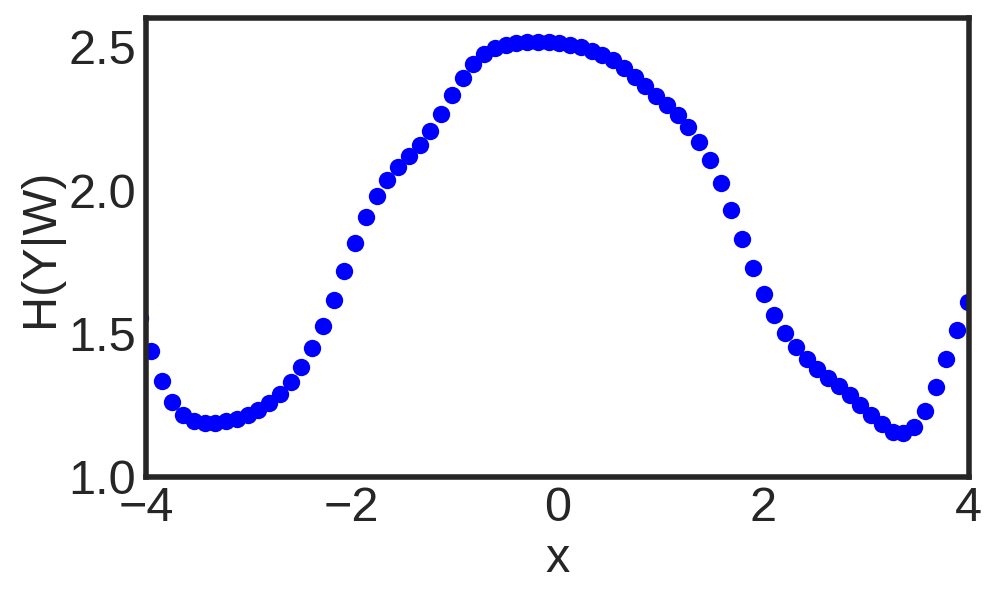}\label{fig:hmchgw}}
\subfloat[][]{\includegraphics[scale=0.21]{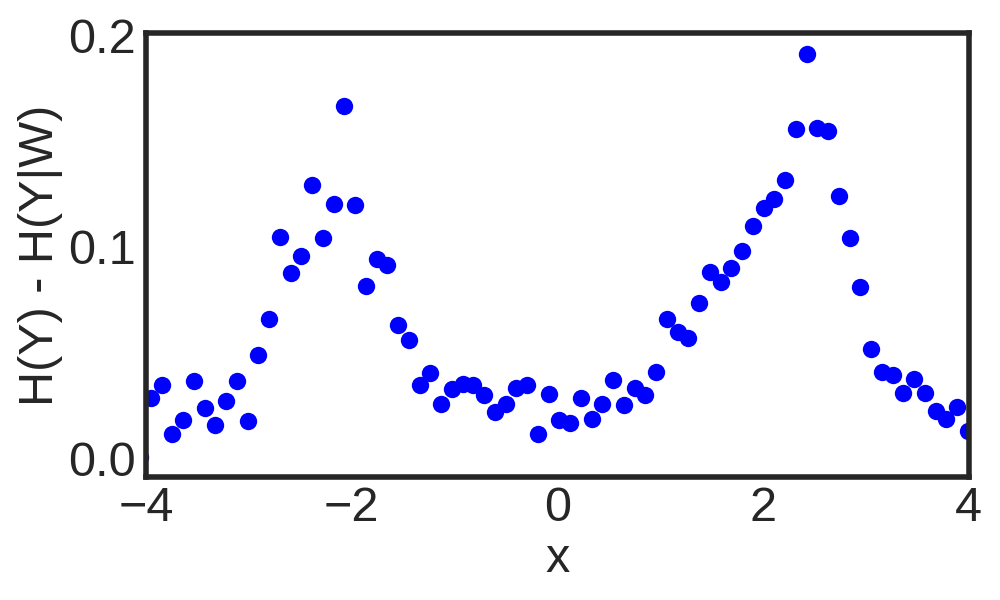}\label{fig:hmcob}}
\caption{HMC results of active learning example using heteroscedastic data. \protect\subref{fig:hmcrd}: Raw data.
  \protect\subref{fig:hmcdd}: Density of $x$ in raw data. \protect\subref{fig:hmcpd}: Predictive distribution: $p(y|x)$
  of BNN using HMC.  \protect\subref{fig:hmch}: Entropy estimate $\text{H}(y|x)$ of predictive distribution for each $x$. \protect\subref{fig:hmchgw}: Conditional Entropy estimate $\mathbf{E}_\mathcal{W} \text{H}(y|x,\mathcal{W})$ of predictive distribution for each $x$. \protect\subref{fig:hmcob}: Estimate of reduction
  in entropy for each $x$.}
  \label{toy_hmc_het}
  \end{figure*}

We define 
the stochastic function
$y= 7 \sin (x) + 3|\cos (x / 2)| \epsilon$ with $\epsilon \sim \mathcal{N}(0,1)$. The data availability is
limited to specific regions of $x$. In particular, we sample 750 values of $x$
from a mixture of three Gaussians with mean parameters 
$\{\mu_1= -4,\mu_2= 0,\mu_3= 4\}$, variance
parameters $\{\sigma_1=\frac{2}{5},\sigma_2=0.9,\sigma_3=\frac{2}{5}\}$ and with each Gaussian component
having weight equal to $1/3$ in the mixture.
Figure \ref{fig:hmcrd} shows the raw data. We have
lots of points at both borders of the $x$ axis and in the center, but little data available in
between. 
 
\subsection{Bimodal Problem}
We consider a toy problem given by a regression task with bimodal data.
We define $x\in[-0.5, 2]$ and $y=10\sin (x)+\epsilon$
with probability $0.5$ and $y=10\cos (x)+\epsilon$, otherwise, where $\epsilon
\sim \mathcal{N}(0,1)$ and $\epsilon$ is independent of $x$. The data availability is
not uniform in  $x$. In particular we sample 750 values of  $x$ from an exponential distribution with $\lambda=2$
\begin{figure*}[h!]
\centering
\subfloat[][]{\includegraphics[scale=0.21]{figures/plots_bimodal/bimodal_1.png}\label{fig:hmcbrd}}
\subfloat[][]{\includegraphics[scale=0.21]{figures/plots_bimodal/bimodal_1_5.png}\label{fig:hmcbdd}}
\subfloat[][]{\includegraphics[scale=0.21]{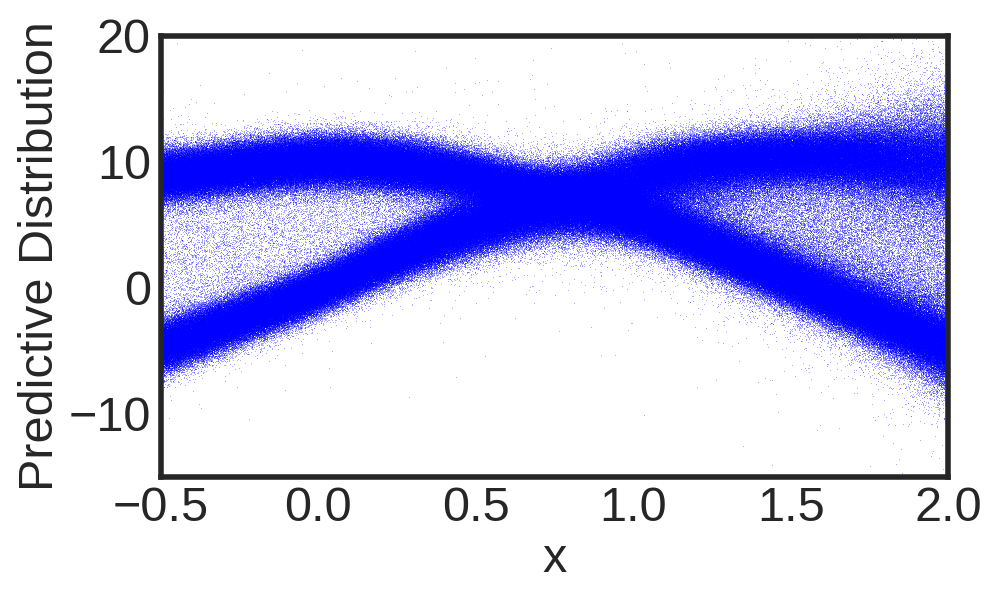}\label{fig:hmcbpd}} \\
\vspace{-0.25cm}
\subfloat[][]{\includegraphics[scale=0.21]{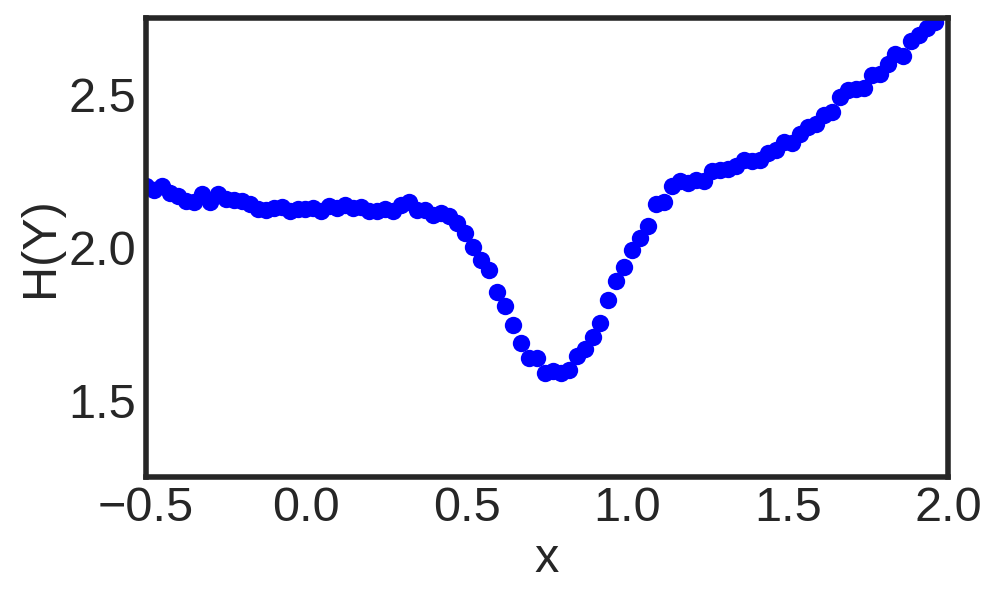}\label{fig:hmcbh}}
\subfloat[][]{\includegraphics[scale=0.21]{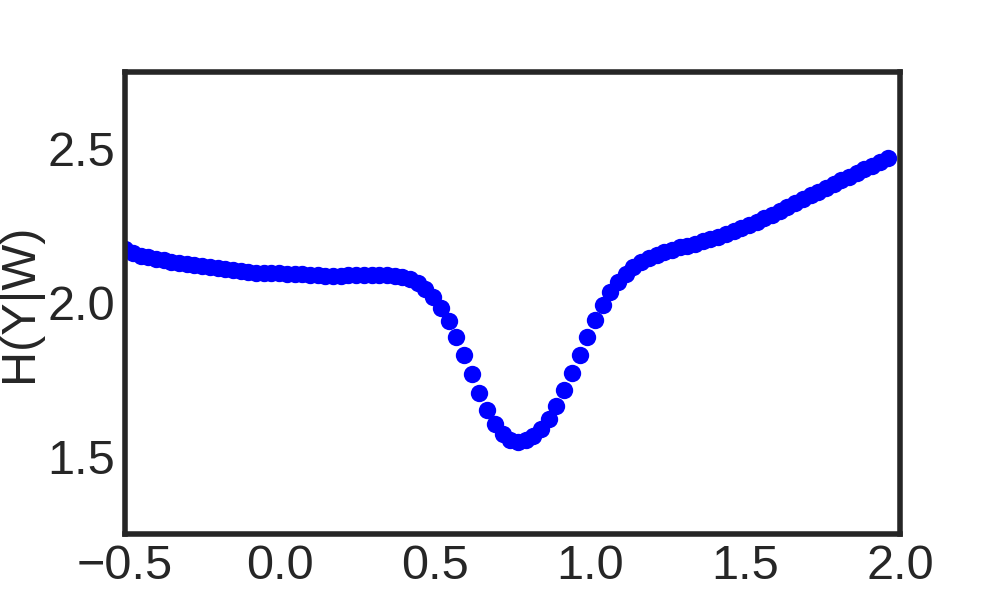}\label{fig:hmcbhgw}}
\subfloat[][]{\includegraphics[scale=0.21]{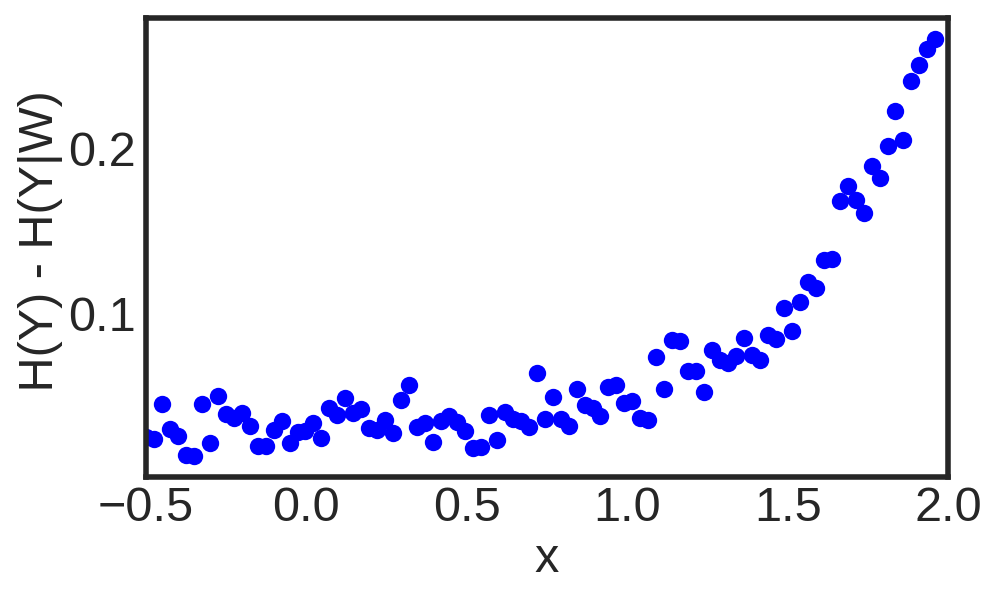}\label{fig:hmcbob}}
\caption{HMC results of active learning example using bimodal data. \protect\subref{fig:hmcbrd}: Raw data.
  \protect\subref{fig:hmcbdd}: Density of $x$ in raw data. \protect\subref{fig:hmcbpd}: Predictive distribution: $p(y|x)$
  of BNN using Hamilton monte carlo.  \protect\subref{fig:hmcbh}: Entropy estimate $\text{H}(y|x)$ of predictive distribution for each $x$. \protect\subref{fig:hmcbhgw}: Conditional Entropy estimate $\mathbf{E}_\mathcal{W} \text{H}(y|x,\mathcal{W})$ of predictive distribution for each $x$. \protect\subref{fig:hmcbob}: Estimate of reduction
  in entropy for each $x$.}
  \label{toy_hmc_bim}
  \end{figure*}

\newpage
\section{Solutions to Toy Problems for different values of $\alpha$}
In the main document we pointed out that the decomposition of uncertainty does not work as good with other values of $\alpha$. We will see in the following, that lower values of $\alpha$ will put more and more emphasis on the latent variable $z$. We observe  that the epistemic uncertainty will vanish as  the $\alpha$-divergence minimization approaches  variational Bayes.

\subsection{$\alpha=0.5$}
\begin{figure*}[h!]
\centering
\subfloat[][]{\includegraphics[scale=0.21]{figures/plots_hetero_new/hetero_1.png}\label{fig:a05rd}}
\subfloat[][]{\includegraphics[scale=0.21]{figures/plots_hetero_new/hetero_1_5.png}\label{fig:a05dd}}
\subfloat[][]{\includegraphics[scale=0.21]{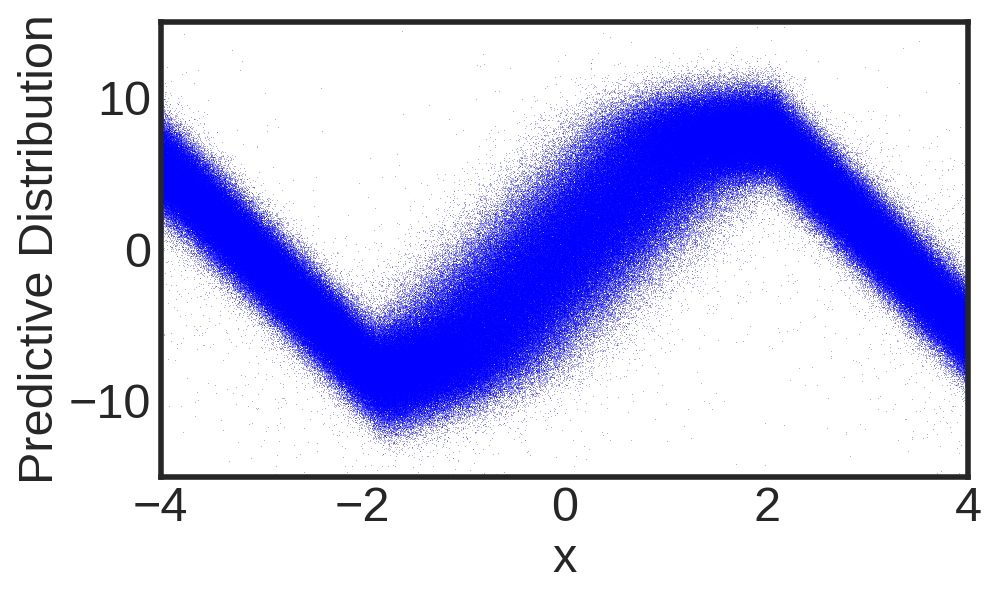}\label{fig:a05pd}} \\
\vspace{-0.25cm}
\subfloat[][]{\includegraphics[scale=0.21]{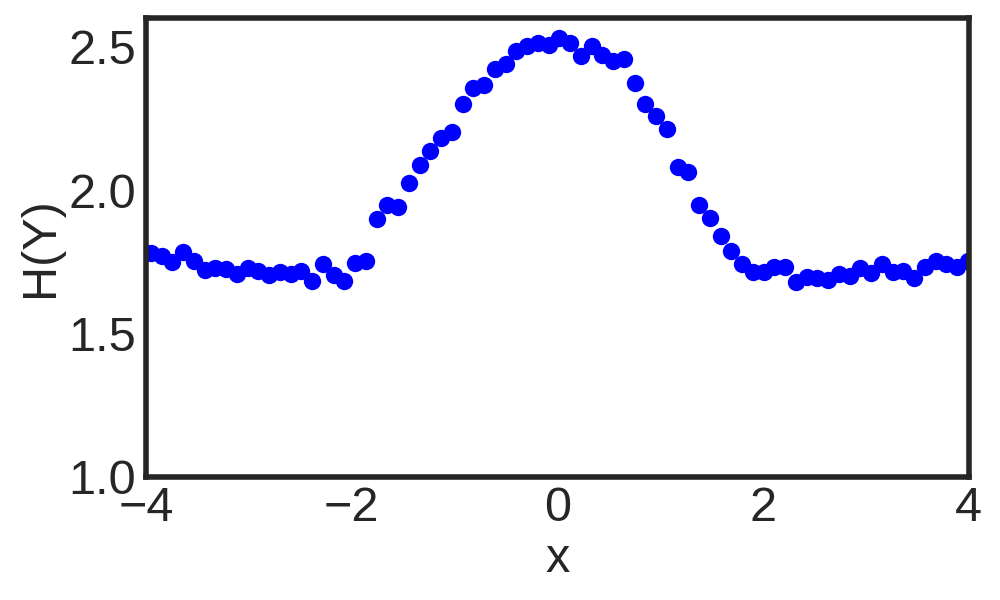}\label{fig:a05h}}
\subfloat[][]{\includegraphics[scale=0.21]{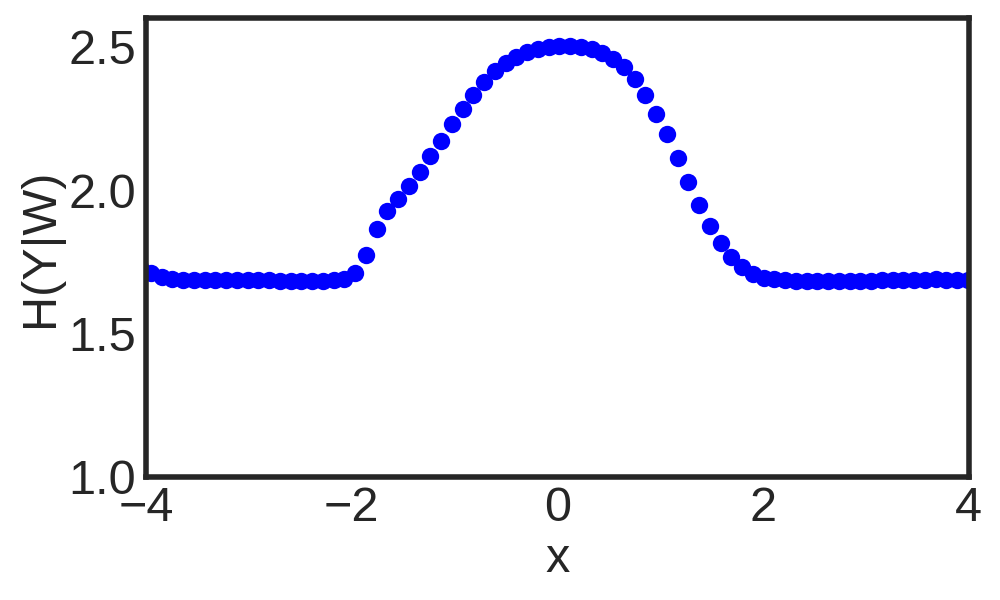}\label{fig:a05hgw}}
\subfloat[][]{\includegraphics[scale=0.21]{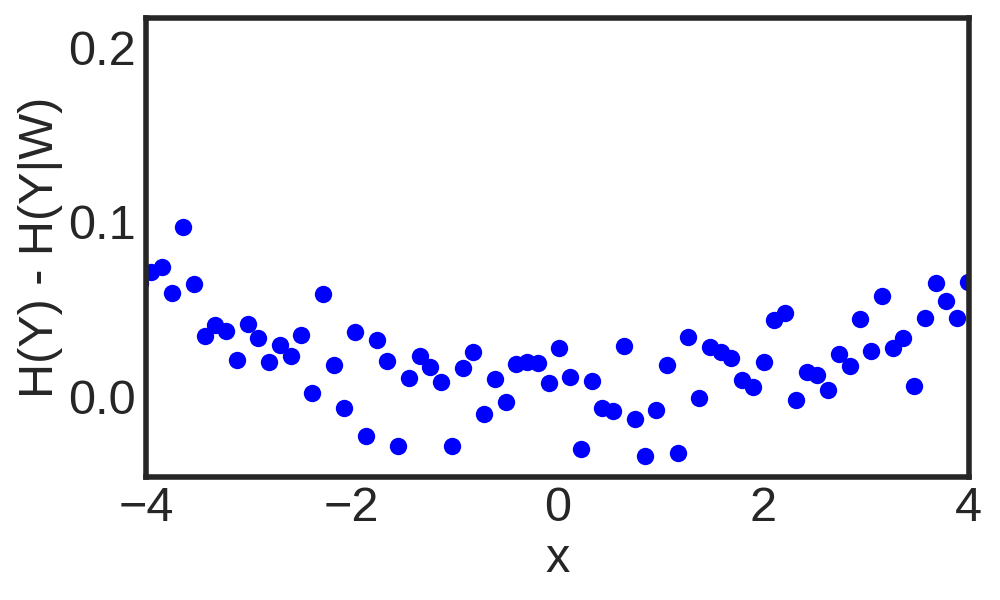}\label{fig:a05ob}}
\caption{Active learning example using heteroscedastic data using a BNN optimized with
bb-$\alpha$  with $\alpha=0.5$. \protect\subref{fig:a05rd}: Raw data.
  \protect\subref{fig:a05dd}: Density of $x$ in raw data. \protect\subref{fig:a05pd}: Predictive distribution: $p(y|x)$
  of BNN.  \protect\subref{fig:a05h}: Entropy estimate $\text{H}(y|x)$ of predictive distribution for each $x$. \protect\subref{fig:a05hgw}: Conditional Entropy estimate $\mathbf{E}_\mathcal{W} \text{H}(y|x,\mathcal{W})$ of predictive distribution for each $x$. \protect\subref{fig:a05ob}: Estimate of reduction
  in entropy for each $x$.}
  \label{toy_a05_het}
  \end{figure*}

\begin{figure*}[h!]
\centering
\subfloat[][]{\includegraphics[scale=0.21]{figures/plots_bimodal/bimodal_1.png}\label{fig:a05brd}}
\subfloat[][]{\includegraphics[scale=0.21]{figures/plots_bimodal/bimodal_1_5.png}\label{fig:a05bdd}}
\subfloat[][]{\includegraphics[scale=0.21]{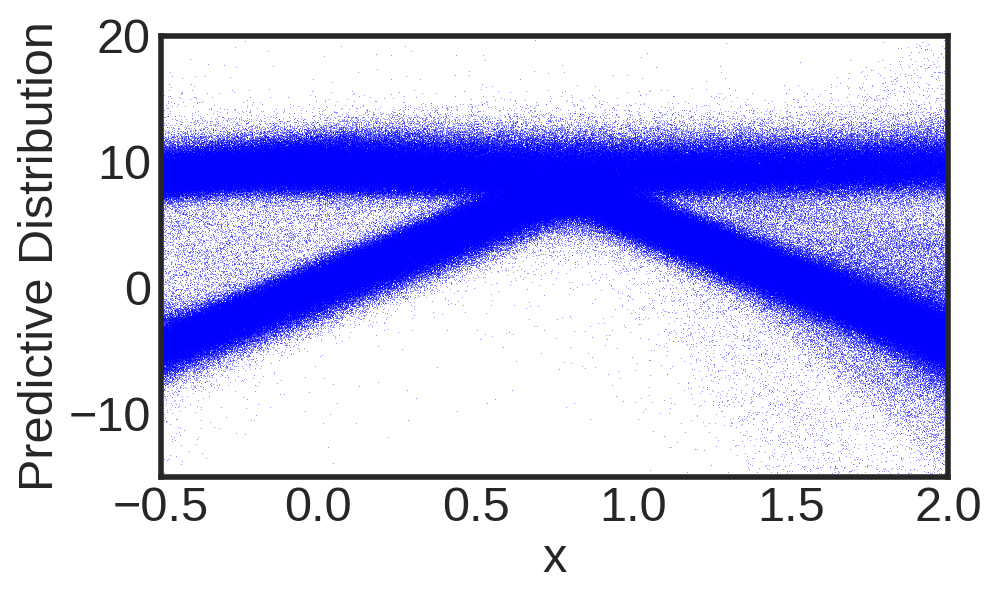}\label{fig:a05bpd}} \\
\vspace{-0.25cm}
\subfloat[][]{\includegraphics[scale=0.21]{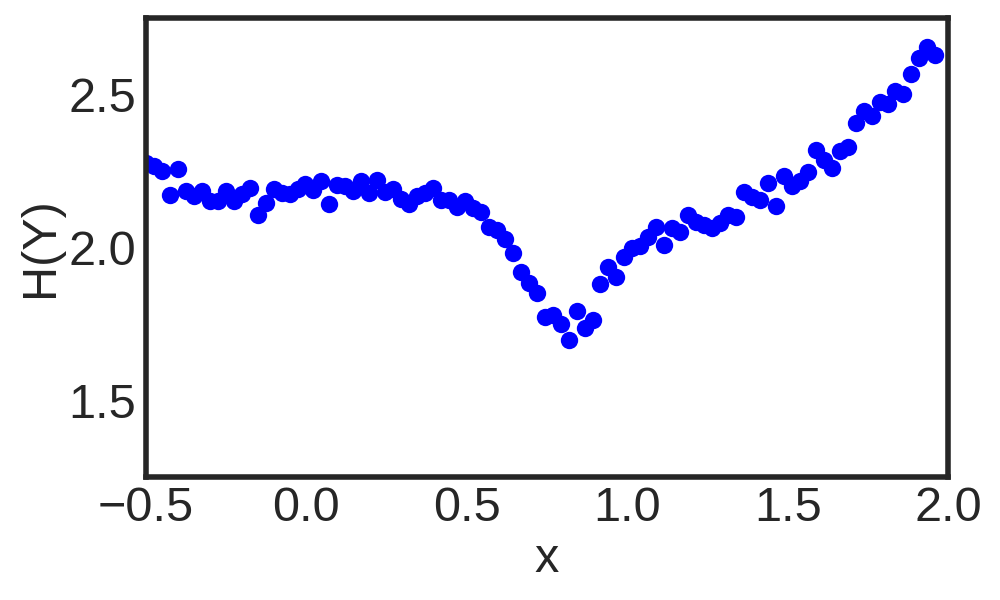}\label{fig:a05bh}}
\subfloat[][]{\includegraphics[scale=0.21]{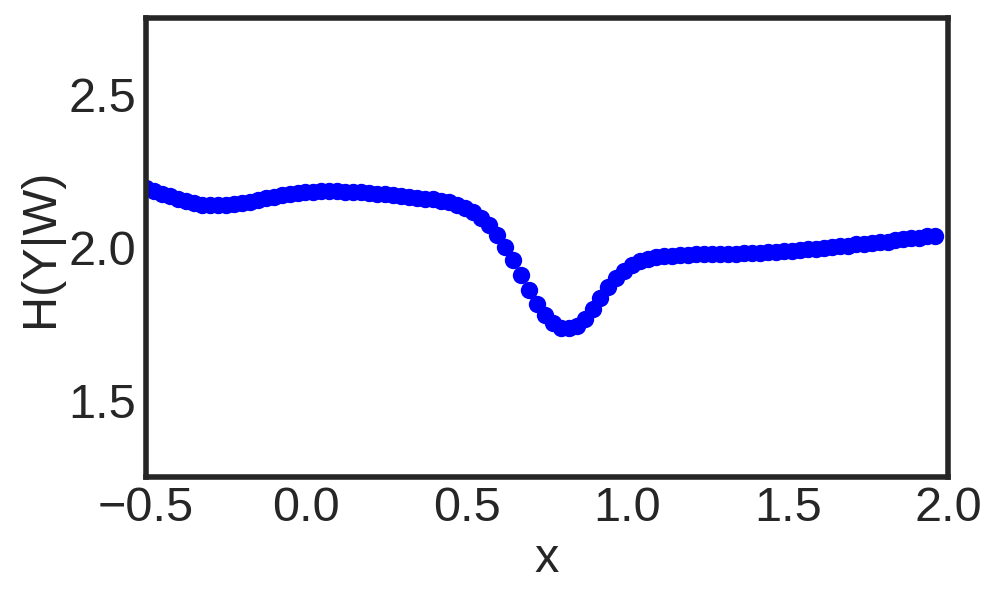}\label{fig:a05bhgw}}
\subfloat[][]{\includegraphics[scale=0.21]{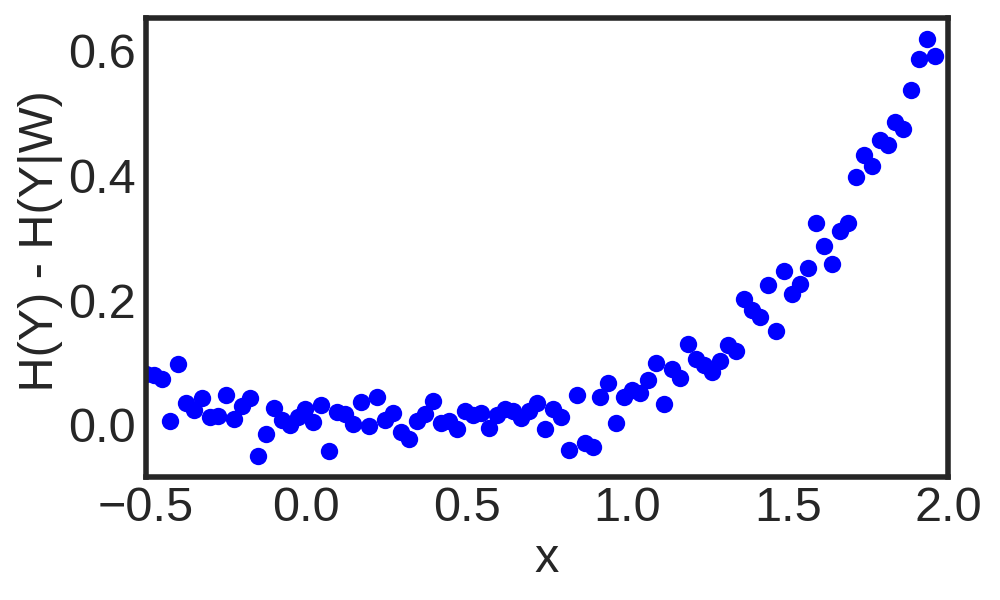}\label{fig:a05bob}}
\caption{Active learning example using bimodal data using a BNN optimized with
bb-$\alpha$  with $\alpha=0.5$. \protect\subref{fig:a05brd}: Raw data.
  \protect\subref{fig:a05bdd}: Density of $x$ in raw data. \protect\subref{fig:a05bpd}: Predictive distribution: $p(y|x)$
  of BNN.  \protect\subref{fig:a05bh}: Entropy estimate $\text{H}(y|x)$ of predictive distribution for each $x$. \protect\subref{fig:a05bhgw}: Conditional Entropy estimate $\mathbf{E}_\mathcal{W} \text{H}(y|x,\mathcal{W})$ of predictive distribution for each $x$. \protect\subref{fig:a05bob}: Estimate of reduction
  in entropy for each $x$.}
  \label{toy_a05_bim}
  \end{figure*}
  
  \newpage
  \subsection{$VB$ solutions to Toy Problems}
  We approximate the method variational Bayes by setting $\alpha$ to $10^{-6}$.
  
\begin{figure*}[h!]
\centering
\subfloat[][]{\includegraphics[scale=0.21]{figures/plots_hetero_new/hetero_1.png}\label{fig:a00rd}}
\subfloat[][]{\includegraphics[scale=0.21]{figures/plots_hetero_new/hetero_1_5.png}\label{fig:a00dd}}
\subfloat[][]{\includegraphics[scale=0.21]{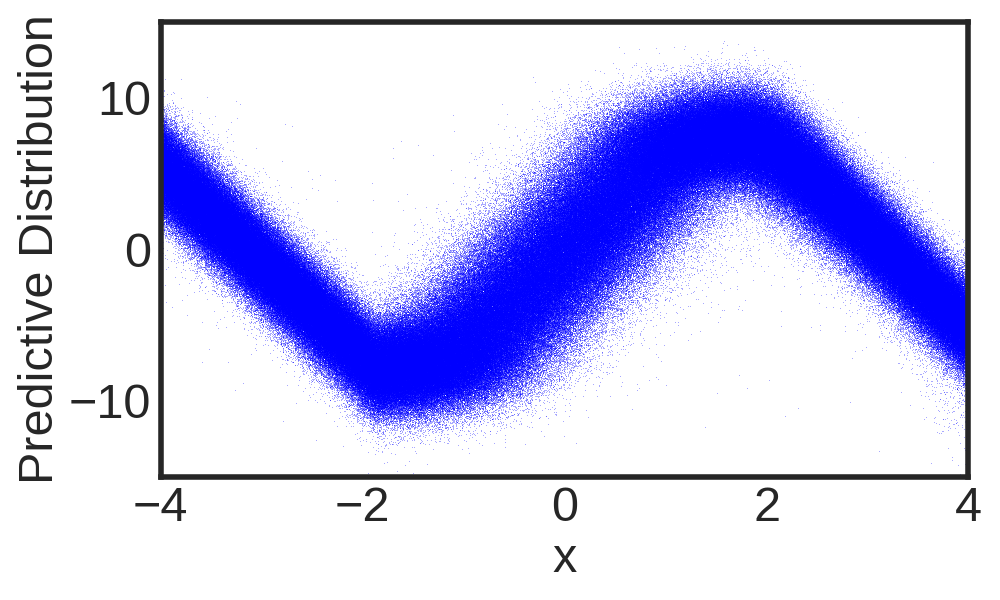}\label{fig:a00pd}} \\
\vspace{-0.25cm}
\subfloat[][]{\includegraphics[scale=0.21]{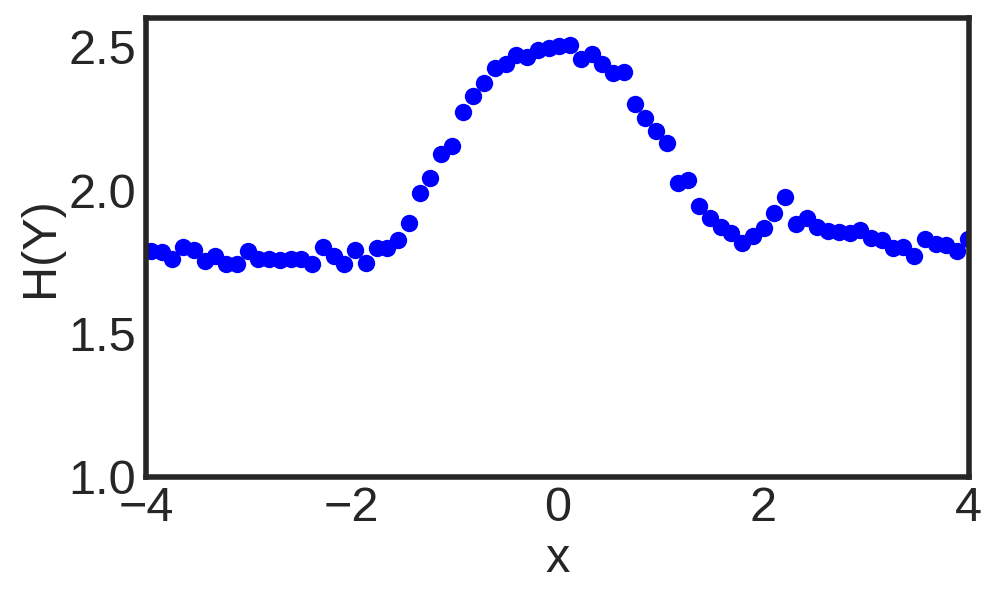}\label{fig:a00h}}
\subfloat[][]{\includegraphics[scale=0.21]{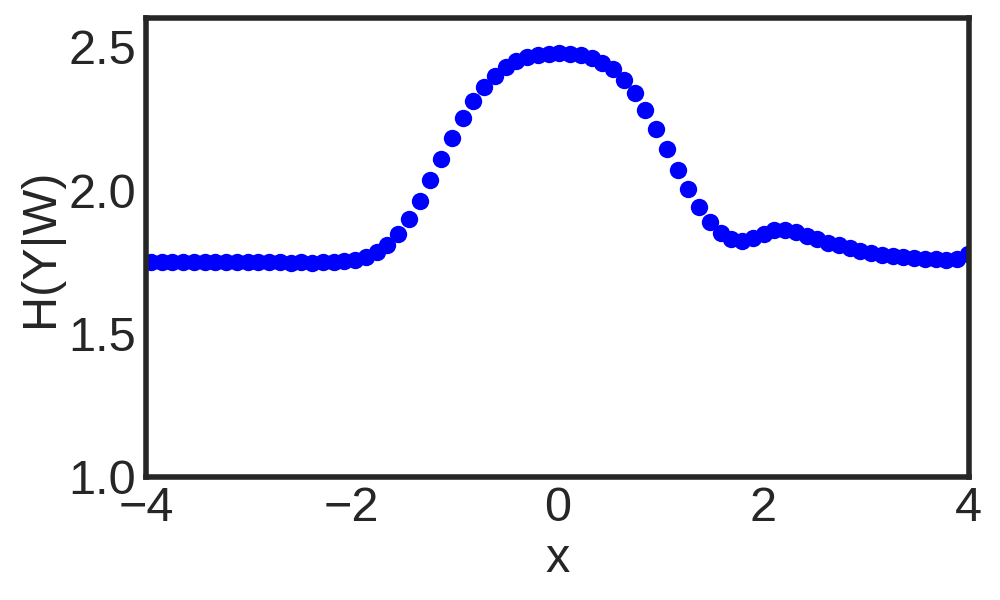}\label{fig:a00hgw}}
\subfloat[][]{\includegraphics[scale=0.21]{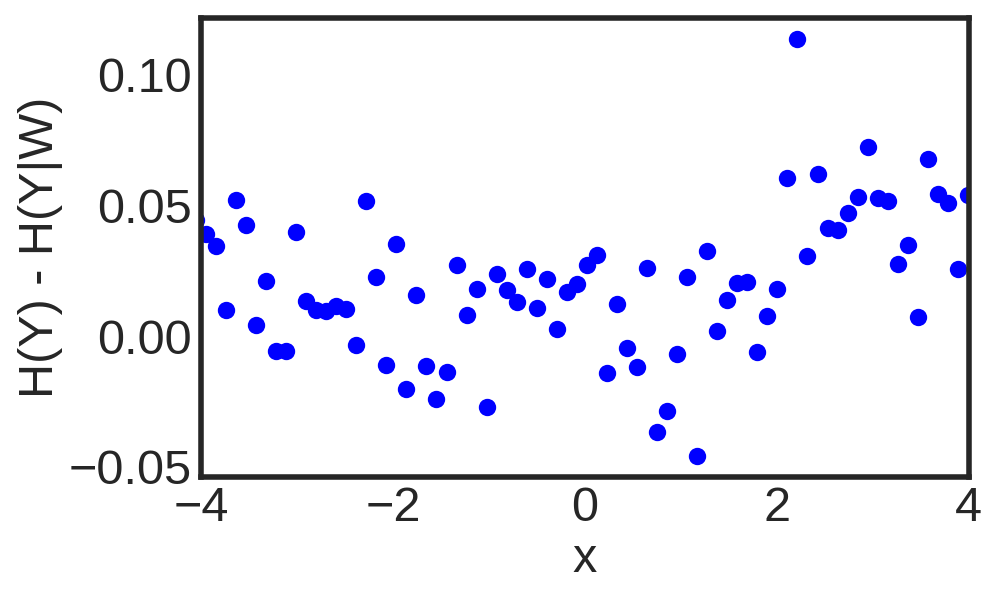}\label{fig:a00ob}}
\caption{Active learning example using heteroscedastic data using a BNN optimized by variational Bayes. \protect\subref{fig:a00rd}: Raw data.
  \protect\subref{fig:a00dd}: Density of $x$ in raw data. \protect\subref{fig:a00pd}: Predictive distribution: $p(y|x)$
  of BNN.  \protect\subref{fig:a00h}: Entropy estimate $\text{H}(y|x)$ of predictive distribution for each $x$. \protect\subref{fig:a00hgw}: Conditional Entropy estimate $\mathbf{E}_\mathcal{W} \text{H}(y|x,\mathcal{W})$ of predictive distribution for each $x$. \protect\subref{fig:a00ob}: Estimate of reduction
  in entropy for each $x$.}
  \label{toy_vb_het}
  \end{figure*}

\begin{figure*}[h!]
\centering
\subfloat[][]{\includegraphics[scale=0.21]{figures/plots_bimodal/bimodal_1.png}\label{fig:a00brd}}
\subfloat[][]{\includegraphics[scale=0.21]{figures/plots_bimodal/bimodal_1_5.png}\label{fig:a00bdd}}
\subfloat[][]{\includegraphics[scale=0.21]{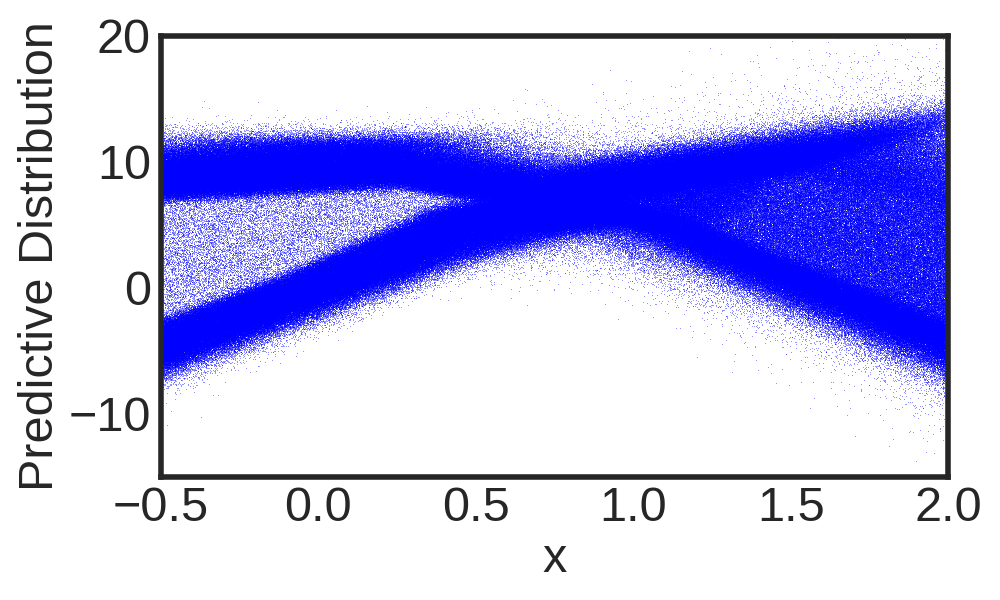}\label{fig:a00bpd}} \\
\vspace{-0.25cm}
\subfloat[][]{\includegraphics[scale=0.21]{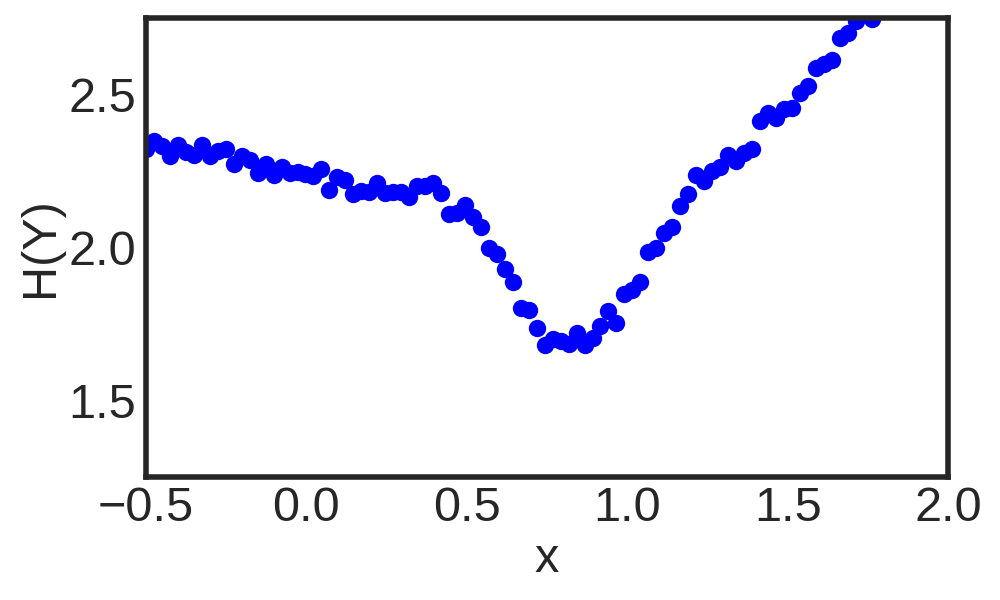}\label{fig:a00bh}}
\subfloat[][]{\includegraphics[scale=0.21]{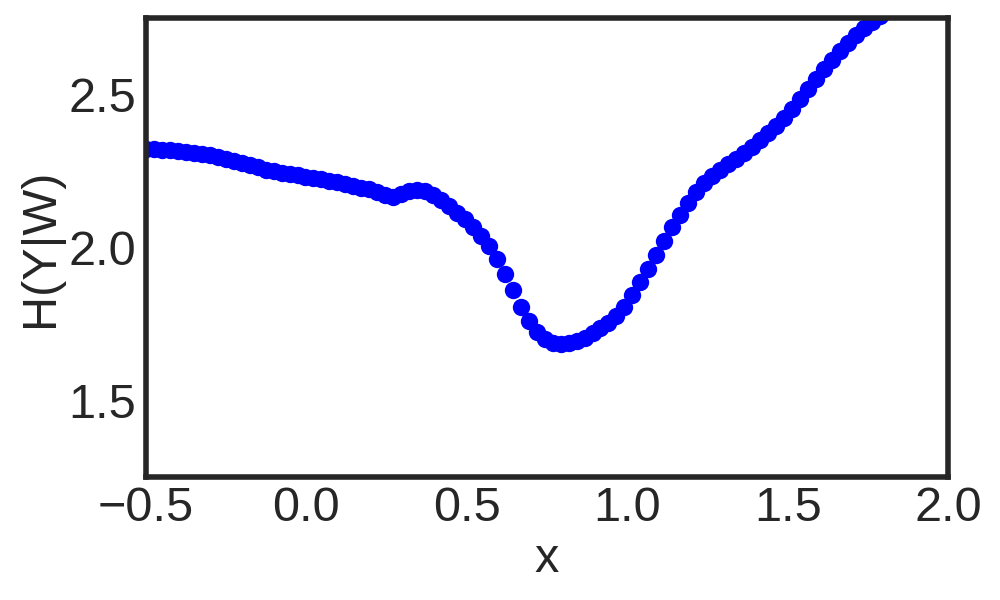}\label{fig:a00bhgw}}
\subfloat[][]{\includegraphics[scale=0.21]{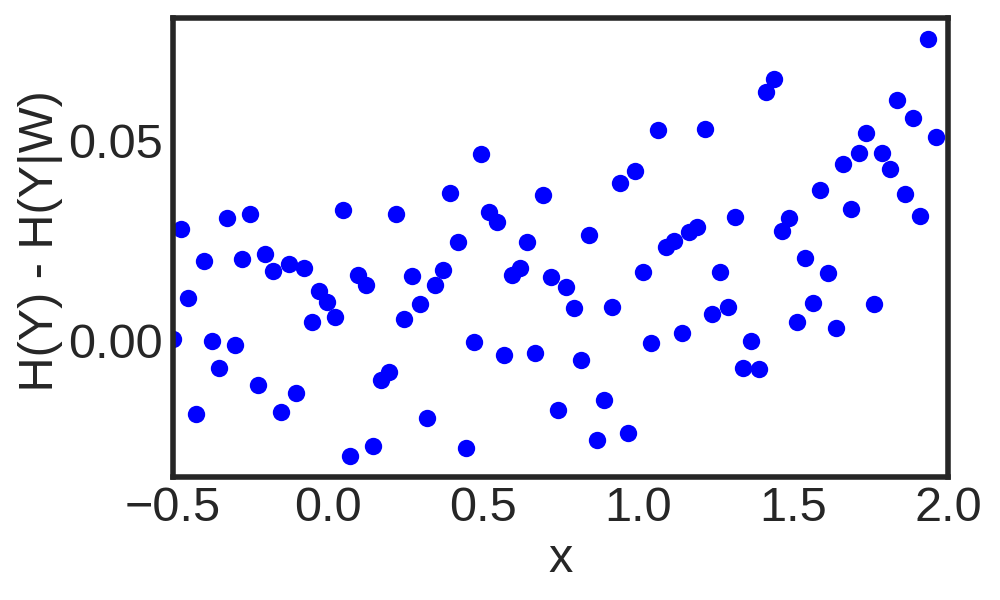}\label{fig:a00bob}}
\caption{Active learning example using bimodal data using a BNN optimized by variational Bayes. \protect\subref{fig:a00brd}: Raw data.
  \protect\subref{fig:a00bdd}: Density of $x$ in raw data. \protect\subref{fig:a00bpd}: Predictive distribution: $p(y|x)$
  of BNN.  \protect\subref{fig:a00bh}: Entropy estimate $\text{H}(y|x)$ of predictive distribution for each $x$. \protect\subref{fig:a00bhgw}: Conditional Entropy estimate $\mathbf{E}_\mathcal{W} \text{H}(y|x,\mathcal{W})$ of predictive distribution for each $x$. \protect\subref{fig:a00bob}: Estimate of reduction
  in entropy for each $x$.}
  \label{toy_vb_bim}
  \end{figure*}
  
 \section{Experiments Specification}
 
 \subsection{Active Learning}
All models start with the available described in the respective paragraphs. We train
for $5000$ epochs a BNN+LV with two-hidden layer and $20$ hidden units per layer. We use
Adam as optimizer with a learning rate of $0.001$.  For Gaussian processes(GPs) we use the
standard RBF kernel using the python GPy implementation.  For the entropy estimation we use a nearest-neighbor
approach as explained in the main document with $k=25$ and $500$ samples of $q(W)$ and $500$ samples of $p(z)$.

For active learning  we evaluate performance using a held-out test set of size $500$ for the bimodal and heteroscedastic problem and $2500$ for the wet-chicken problem. The test data is sampled uniformly in state (and action) space. In each of the $n=150$ iterations we sample a pool set of size $50$ uniformly in input space. In each iteration a method can decide to include $5$ data points into the training set. After that, the models
are re-trained (from scratch) and the performance is evaluated on the test set.


 \subsection{Wet-chicken}
We use the continuous two-dimensional version of the problem.
A canoeist is paddling on a two-dimensional river. The
canoeist's position at time $t$ is $(x_t,y_t)$. The river has width $w=5$ and
length $l=5$ with a waterfall at the end, that is, at $y_t=l$. The canoeist
wants to move as close to the waterfall as possible because at time $t$ he gets
reward $r_t = -(l - y_t)$. However, going beyond the waterfall boundary makes the
canoeist fall down, having to start back again at the origin $(0,0)$. At
time $t$ the canoeist can choose an action $(a_{t,x},a_{t,y}) \in [-1,1]^2$ that represents
the direction and magnitude of his paddling. The river dynamics have stochastic
turbulences $s_t$ and drift $v_t$ that depend on the canoeist's position on the
$x$ axis. The larger $x_t$, the larger the drift and the smaller $x_t$, the
larger the turbulences.

The underlying dynamics are given by the following system of equations. The
drift and the turbulence magnitude are given by $v_t=3x_tw^{-1}$ and $s_t = 3.5
- v_t$, respectively. The new location $(x_{t+1},y_{t+1})$ is given
by the current location $(x_t,y_t)$ and current action $(a_{t,x},a_{t,y})$ using
\begin{align}
x_{t+1} &= \begin{cases} 
0 & \text{if } \quad x_t + a_{t,x} < 0 \\
0 & \text{if } \quad \hat{y}_{t+1} > l \\
w & \text{if }  \quad x_t + a_{t,x} > w \\
x_t + a_{t,x} & \text{otherwise}  \end{cases}\,,   
&
y_{t+1} &= \begin{cases}
0 & \text{if } \quad \hat{y}_{t+1} < 0 \\
0 & \text{if } \quad \hat{y}_{t+1} > l \\
\hat{y}_{t+1} & \text{otherwise} \end{cases}\,,
\end{align}
where $\hat{y}_{t+1} =  y_t + (a_{t,y} - 1) + v_t + s_t \tau_t$
and $\tau_t \sim \text{Unif}([-1,1])$ is a random variable that represents the
current turbulence.
 As the canoeist moves closer to the waterfall, the
distribution for the next state becomes increasingly bi-modal because when he is close to the waterfall, the change in
the current location can be large if the canoeist falls down the waterfall and
starts again at $(0,0)$. The distribution may also be truncated uniform for states close to
the borders.  Furthermore the system has
heteroscedastic noise, the smaller
the value of $x_t$ the higher the noise variance.
 \subsection{Reinforcement Learning}
 
 \subsubsection{Industrial benchmark}

 Policies in the industrial benchmark specify
changes $\Delta_v$, $\Delta_g$ and $\Delta_s$ in three steering variables $v$ (velocity), $g$ (gain)
and $s$ (shift) as a function of $\mathbf{s}_t$. In the behavior
policy these changes are stochastic and sampled according to
\begin{align}
\Delta_v & =  \left\{
\begin{array}{@{}ll@{}}
   z_v\,, & \text{if}\,\, v(t) < 40 \\
    -z_v\,, & \text{if}\,\, v(t) > 60 \\
    u_v \,, & \text{otherwise}
  \end{array}
  \right.\\
\Delta_g & = \left\{
\begin{array}{@{}ll@{}}
    z_g\,, & \text{if}\,\, g(t) < 40 \\
    -z_g\,, & \text{if}\,\, g(t) > 60 \\
    u_g \,, & \text{otherwise}
  \end{array}
  \right.\\
\Delta_s & =  u_s \,,
\end{align}
where $z_v,z_g \sim \mathcal{N}(0.5,\frac{1}{\sqrt{3}})$ and
$u_v,u_g,u_s\sim \mathcal{U}(-1,1)$.  The velocity $v(t)$ and gain
$g(t)$ can take values in $[0,100]$. Therefore, the data collection
policy will try to keep these values only in the medium range given by
the interval $[40,60]$. Because of this, large parts of the state
space will be unobserved. After collecting the data, the $30,000$
state transitions are used to train a BNN with latent variables with
the same hyperparameters as in \cite{depeweg2016learning}.  Finally,
we train different policies using the Monte Carlo approximation
and we set the horizon of
$T=100$ steps, with $M=50$ and $N=25$ and a minibatch size of $1$ for $750$ epochs. The total
training time on a single CPU is around $18$ hours.

\subsubsection{Wind Turbine Simulator}
In this problem we observe the turbine state $s(t)$, with features such as  current wind speed and currently produced power. Our actions $a(t)$ adjust the turbine's behavior, with known upper and lower bounds. The goal is to maximize energy output over a $T$-step horizon.

We are given a batch of around 5,000 state transitions generated by a behavior policy $\pi_b$.  The policy 
does limited exploration around the neutral action $a(t)=0$.

The system is expected to be highly stochastic due to the unpredictability of future wind dynamics. Furthermore the dimensionality of state observation is much higher than the action dimensionality, so, with the limited dataset that we have, we  expect it to be very challenging to accurately learn the influence of the action on the reward.

First we train a BNN with two hidden layer and 50 hidden units per layer on the available batch using $\alpha$-divergence minimization with $\alpha=1.0$. In the second step, using the model, we train a policy with 20 hidden units on each of the two layers in the usual way, using the Monte Carlo estimate. The total
training time on a single CPU is around $8$ hours.

\end{document}